\documentclass{fairmeta}
\usepackage[utf8]{inputenc}
\usepackage{amsmath,amssymb,amsfonts,amsthm,mathtools}
\usepackage{tabularx,makecell,adjustbox}
\usepackage[ruled,vlined]{algorithm2e}
\usepackage{fancyhdr}
\usepackage{array}
\usepackage{multirow}
\usepackage{url}
\usepackage{verbatim}
\usepackage{enumitem}

\makeatletter
\def\input@path{{B-migrated/}}
\makeatother
\graphicspath{{B-migrated/}}
\fancypagestyle{gammaTeleAI}{%
  \fancyhf{}
  \fancyfoot[C]{\thepage}
  \renewcommand{\headrulewidth}{\ifnum\value{page}>1 0.3pt\else 0pt\fi}
  
}
\makeatletter
\let\ps@plain\ps@gammaTeleAI
\makeatother

\title{KODA: Kinematics--Observation Data Alignment for Humanoid Learning from Human Egocentric Demonstration}
\author[1,2,3,\dagger]{\text{Zibo Zhou}}
\author[2,3,\dagger]{\text{Yueru Chen}}
\author[2,3]{\text{Junsong Wu}}
\author[2,3]{\text{Weiji Xie}}
\author[3]{\text{Qingyao Xu}}
\author[3]{\text{Sheng Yin}}
\author[2,4]{\text{Shiqing Wang}}
\author[1,2]{\text{Jiyuan Shi}}
\author[3]{\text{Siheng Chen}}
\author[1,2,*]{\text{Chenjia Bai}}
\author[2,*]{\text{Xuelong Li}}
\affiliation[1]{Gamma Robotics}
\affiliation[2]{Institute of Artificial Intelligence (TeleAI), China Telecom }
\affiliation[3]{Shanghai Jiao Tong University}
\affiliation[4]{Zhejiang University}
\affiliation[\dagger]{Equal contribution (co-first authors)}
\affiliation[*]{Corresponding authors}
\date{\today}
\project{\url{https://koda-humanoid.github.io/koda/}}
\metadata[Correspondence to]{Chenjia Bai (\email{baicj@chinatelecom.cn})\\ \ }
\abstract{%
Training humanoid whole-body vision--language--action (VLA) policies requires
costly robot demonstrations with aligned robot-view observations and
controller-compatible whole-body state--action labels. Egocentric human
demonstrations offer scalable supervision but have misaligned views and
sparse motion cues. Existing approaches either pretrain policies that still
require robot data for adaptation or generate controller-specific commands
that limit reuse. Conversion also requires robot-view observations aligned
with the generated whole-body motion. We present KODA
(Kinematics--Observation Data Alignment), which converts egocentric RGB
demonstrations into whole-body VLA training data using a reusable canonical
trajectory to guide robot-view synthesis and generate controller-specific
whole-body state--action supervision. KODA first generates a
simulator-checked whole-body trajectory using a sparse tracker with
sampling-based command optimization. This generated trajectory directly
determines the onboard camera poses used to synthesize corresponding
robot-view observations. The motion, observations, and task instruction form
a canonical record. For each deployment controller, KODA optimizes commands
to track the stored motion and records the corresponding action labels and
proprioceptive states. When switching controllers, KODA retains this
canonical record and regenerates whole-body state--action supervision for
co-training with matching robot demonstrations. We evaluate KODA on four
Unitree G1 loco-manipulation tasks using two whole-body controllers, SONIC
and HEFT. With robot data held fixed, macro-average success on conditions
covered by human demonstrations but absent from robot data improves by
45.0 and 46.25 percentage points, respectively. Under both controllers, this
metric shows an upward trend across the evaluated human-data budgets,
suggesting potential for further scaling.\par\medskip
\textbf{Keywords:} Egocentric Human Data, Humanoid Loco-Manipulation}

\begin{document}
\maketitle

\begin{figure}[!t]
  \centering
  \includegraphics[width=0.9\textwidth]{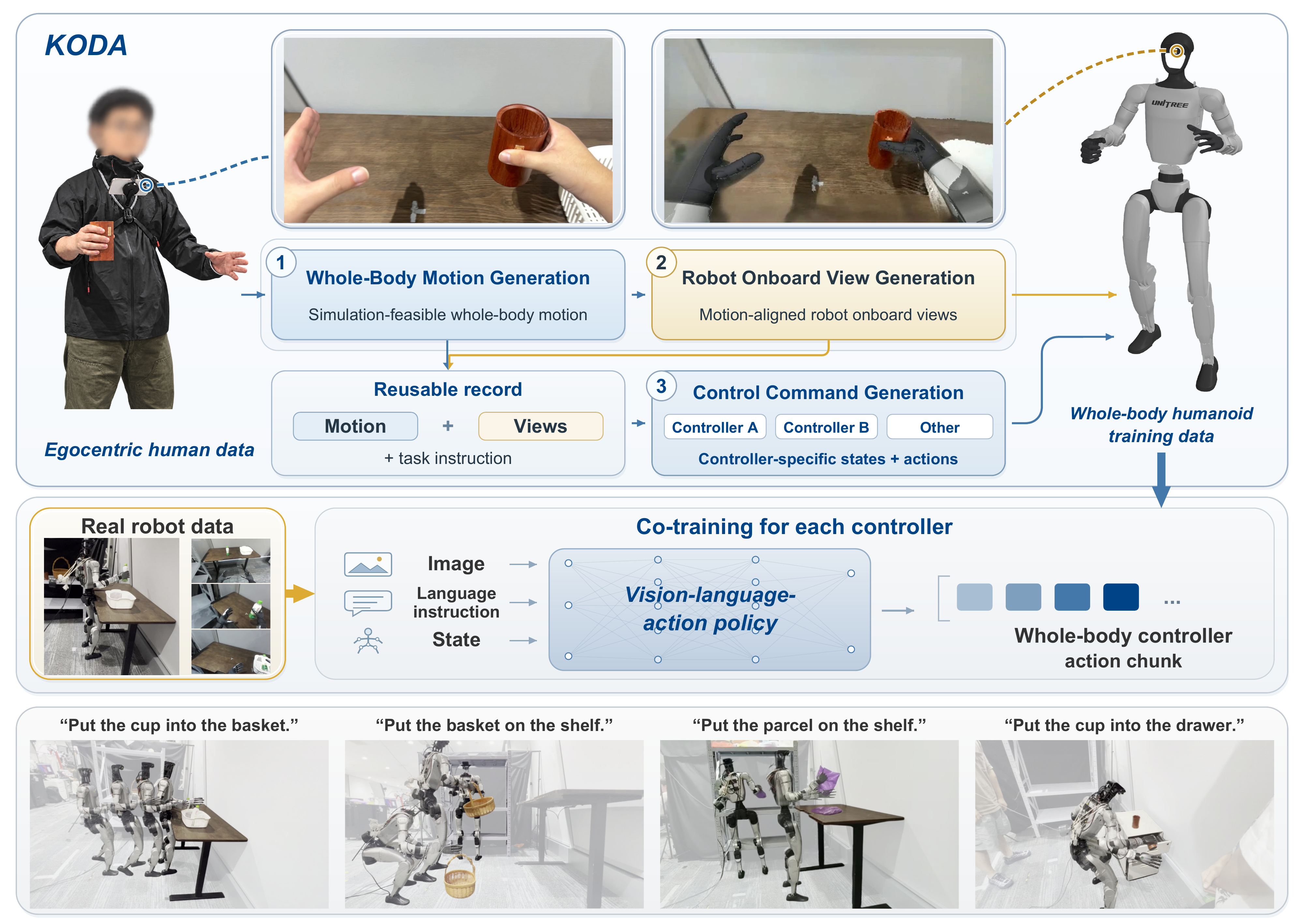}
  \caption{Overview of KODA. \textbf{Whole-Body Motion Generation} produces simulation-feasible whole-body motion from human egocentric demonstrations, and \textbf{Robot Onboard View Generation} synthesizes motion-aligned onboard views. The motion, views, and task instruction form a reusable canonical record. \textbf{Control Command Generation} uses this record to generate controller-specific states and actions. The resulting data are combined with matching real-robot demonstrations to co-train a controller-specific vision--language--action (VLA) policy for language-guided humanoid loco-manipulation.}
  \label{fig:cem-inner}
\end{figure}

\section{Introduction}

Humanoid robots combine locomotion and dexterous manipulation in a
versatile platform for operating in environments designed for people.
Realizing this potential requires policies that coordinate body posture,
mobility, and bimanual manipulation across the whole body.
Whole-body vision--language--action (VLA) models aim to learn such
coordination from onboard observations and task
instructions~\citep{jiang2026wholebodyvla,hu2026openhlm}. Training
requires corresponding robot observations, whole-body states, and action
supervision compatible with the deployment whole-body controller (WBC)
interface. Collecting these demonstrations on the target humanoid is
costly, limiting coverage across tasks, objects, and environments.

Human demonstration interfaces and egocentric datasets provide RGB images,
camera poses, and sparse bilateral hand-position references at
scale~\citep{chi2024universal,hoque2026egodex,punamiya2026egoverse}.
Their success on fixed-base arms~\citep{chi2024universal} and in dexterous
manipulation~\citep{zheng2026egoscale,liu2026egoengine} motivates extending
human-data supervision to whole-body VLAs.
Approaches such as EgoVLA, WholeBodyVLA, and MotionWAM learn
human-derived priors for humanoid policies but still depend on
target-humanoid demonstrations for adaptation~\citep{yang2025egovla,jiang2026wholebodyvla,zheng2026motionwam}.
Target-humanoid data collection therefore remains a bottleneck.
Other work converts human demonstrations into robot-oriented supervision.
EgoHumanoid uses whole-body human measurements and a shared
end-effector-and-navigation action space for co-training~\citep{shi2026egohumanoid}.
HALOMI trains policies on human demonstrations using head--hand command
labels tied to a dedicated controller~\citep{zhao2026halomi}. These methods
produce supervision tied to particular action interfaces, creating data
silos that limit reuse across whole-body controllers; their outputs do not directly
provide the whole-body states and compatible actions needed for other WBC
interfaces. Their converted episodes also require observations corresponding
to the generated robot motion. Fixed-offset reprojection~\citep{shi2026egohumanoid}
shifts the human camera without following the generated robot trajectory,
leaving this observation--motion correspondence unresolved. Moreover,
retaining only controller-tuned commands limits reuse to that controller's
command space and tracking behavior. A reusable record must support
adaptation to different deployment controllers.

We target a distinct data-conversion setting: augmenting an existing
controller-specific humanoid whole-body dataset from egocentric RGB, camera
poses, and hand references. This setting requires three capabilities:
(i) generating simulator-checked whole-body trajectories from sparse
human references;
(ii) synthesizing onboard observations aligned with the generated robot
motion; and
(iii) instantiating the same converted record as compatible state--action
supervision for different deployment controllers.

To address these three challenges, we present KODA
(Kinematics--Observation Data Alignment), a framework that converts
egocentric RGB demonstrations into whole-body VLA training data.
The conversion proceeds in three stages, as shown in Fig.~\ref{fig:cem-inner}.
First, KODA generates a whole-body trajectory from sparse human references
using a fine-tuned tracker. Sampling-based command optimization improves
reference tracking while maintaining trajectory feasibility in simulation.
Second, KODA derives camera poses from the generated trajectory and
synthesizes corresponding onboard observations. The aligned motion and
observations, together with the task instruction, form a canonical record
for the target humanoid.
Third, KODA optimizes the selected deployment controller's commands to
track the fixed canonical trajectory, producing compatible state--action
supervision for co-training with matching real-robot demonstrations.
Changing controllers regenerates the state--action supervision while
retaining the canonical motion and robot-view observations.

Our contributions are:
(a) A data-conversion framework that constructs whole-body robot motion
and corresponding onboard observations from egocentric human demonstrations.
(b) A canonical representation and controller-specific control command generation
procedure for generating compatible state--action supervision from the
same converted records across deployment interfaces.
(c) Real-world evidence of the training value and scaling potential of
converted human demonstrations on four G1 loco-manipulation tasks.
With 50 robot demonstrations per task fixed, adding 100 converted human
demonstrations per task increases macro-average success by 45.0 and 46.25
percentage points under SONIC and HEFT, respectively, on object
conditions covered by human training clips but absent from robot
demonstrations. This metric shows an upward trend across the evaluated data budgets
under both interfaces.

\begin{figure}[!t]
  \centering
  \includegraphics[width=1.0\textwidth]{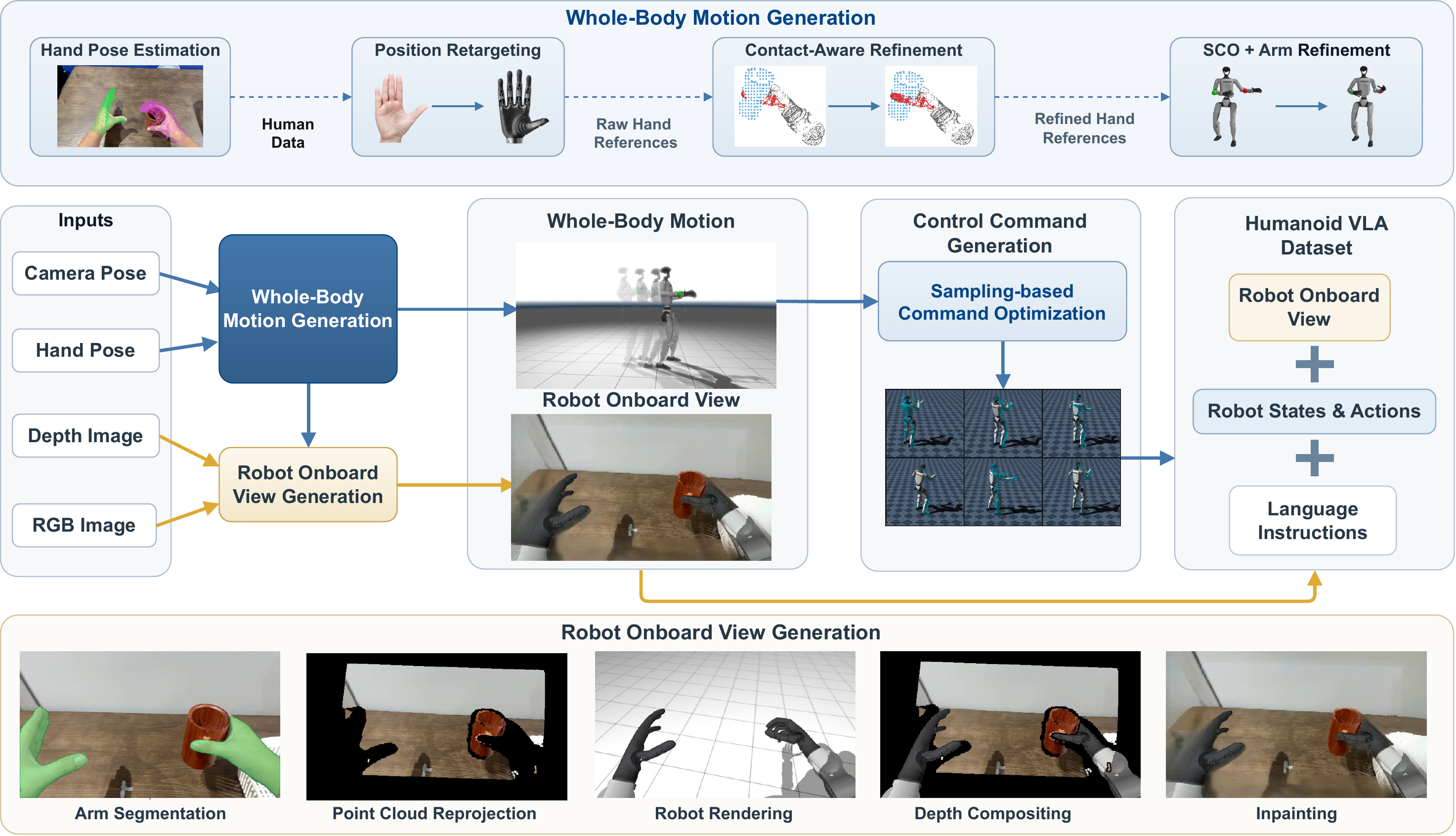}
  \caption{KODA pipeline. \textbf{Middle:} Human demonstrations become aligned whole-body motion and onboard observations, then controller-specific state--action data for VLA training (Control Command Generation).
  \textbf{Top:} Whole-Body Motion Generation combines hand estimation and contact-aware retargeting, sampling-based command optimization (SCO), and arm refinement.
  \textbf{Bottom:} Robot Onboard View Generation combines arm removal, scene reprojection, robot rendering, depth compositing, and inpainting. Input depth is observed or estimated from RGB.}
  \label{fig:koda-overview}
\end{figure}
\section{Related Work}
\subsection{Humanoid Loco-Manipulation}
Humanoid data collection methods use direct teleoperation, human imitation, or
whole-body tracking to obtain robot demonstrations. OmniH2O, HOMIE, HumanPlus,
and TWIST2 provide interfaces for whole-body teleoperation or imitation
learning~\citep{he2024omnih2o,ben2025homie,fu2024humanplus,ze2025twist2}.
OmniRetarget preserves human--object and human--environment interactions during
retargeting, while SONIC tracks whole-body motion for physically grounded
control~\citep{yang2025omniretarget,luo2026sonic,xie2025kungfubot,han2026kungfubot2}. WholeBodyVLA learns
loco-manipulation from vision and instructions, while OpenHLM and MotionWAM
study whole-body humanoid policy and action-model training
\citep{jiang2026wholebodyvla,hu2026openhlm,zheng2026motionwam,xie2026textop}. These approaches
establish the control and learning foundation for humanoid loco-manipulation,
but their supervision is collected or represented in a target-robot interface.

\subsection{Egocentric Data and Cross-Embodiment Alignment}
Egocentric interfaces and datasets provide a scalable source of visual and
motion supervision. UMI records robot-compatible gripper trajectories and
images, while EgoDex and EgoVerse provide large-scale RGB data with camera and
hand-motion annotations~\citep{chi2024universal,hoque2026egodex,punamiya2026egoverse}.
$\Psi_0$ and EgoVLA use human data for embodied pretraining, and EMMA,
EgoEngine, and Ego2Robot combine human video with robot learning or robot-data
synthesis~\citep{wei2026psi0,yang2025egovla,zhu2026emma,liu2026egoengine,wang2026ego2robot}.
Related cross-embodiment methods address appearance and viewpoint transfer
through cross-painting or robot-view augmentation~\citep{chen2024mirage,chen2024rovi}.

For humanoids, EgoHumanoid uses whole-body measurements, fixed-offset view
alignment, and training-time perturbations~\citep{shi2026egohumanoid}.
HALOMI adopts this view alignment and reference adaptation based on the cross-entropy method (CEM) for a
dedicated controller~\citep{zhao2026halomi}. KODA also uses command optimization,
but derives onboard views from generated whole-body motion. Its canonical
motion--observation records can be reused to augment manipulation datasets
for compatible WBCs through controller-specific command generation and
state recording.

\section{System Framework}
\label{sec:method}

KODA converts egocentric human episodes into robot-view observations, canonical
whole-body motion, and controller-compatible state--action data for VLA training
as shown in Fig.~\ref{fig:koda-overview}. The conversion has three stages:
Whole-Body Motion Generation in Sec.~\ref{sec:kinematic_generation} produces a
simulator-checked trajectory from sparse human references; Robot Onboard View
Generation in Sec.~\ref{sec:robot_view_generation} uses the corresponding
body--camera trajectory to synthesize aligned observations, forming a
controller-independent canonical record; and Control Command Generation in
Sec.~\ref{sec:vla_training} tracks this motion with the selected controller
to derive its state--action supervision. The canonical motion and observations
are reused across interfaces, while controller-specific states and commands
enable co-training with matching real-robot demonstrations.
Qualitative demonstration timelines are provided in Appendix~\ref{app:qualitative}.

\subsection{Problem Setup}\label{sec:problem_setup}
Let $\mathcal R$ denote the target humanoid and $\kappa$ a deployment
controller with its action interface. All sequences use the target control
clock $t=1,\ldots,T$.

After geometry recovery, an RGB episode is represented as
$e_h=(\{I_t^h,D_t^h,C_t^h,J_t^h\}_{t=1}^{T},\ell)$, where $I_t^h$ is
the egocentric RGB image, $D_t^h$ is observed or estimated depth,
$C_t^h$ is the source-camera pose,
$J_t^h$ contains the bilateral 3-D hand keypoints, and $\ell$ is the
task instruction. Camera poses and hand keypoints come from source
annotations or offline recovery. When source depth is unavailable, KODA
estimates it from RGB using the metric-depth model of Depth Anything
3~\citep{lin2025depthanything3}. Depth, camera poses, and hand references
must share a consistent metric coordinate system.
Human episodes form $\mathcal D_{\mathrm{human}}$.
Real-robot episodes in $\mathcal D_{\mathrm{robot}}^\kappa$ contain
robot-view observations $I_t^{\mathcal R}$, controller-specific state
features $s_t^\kappa$, actions $a_t^\kappa$, and instructions $\ell$.

$\mathcal C_{\mathcal R}$ constructs a canonical record in $\mathcal R$'s observation
and kinematic spaces, without controller-specific states or actions:
$e_{\mathcal R}^{\mathrm{can}}=\mathcal C_{\mathcal R}(e_h)=
\left(\left\{I_t^{\mathcal R},q_t^{\mathcal R}\right\}_{t=1}^{T},\ell\right)$.
Here $q_t^{\mathcal R}$ is the whole-body configuration corresponding to
$I_t^{\mathcal R}$, including the floating-base pose and body and hand
joint positions.
The canonical trajectory provides a common motion reference for robot-view
synthesis and controller-specific tracking.

The mappings $\mathcal C_{\mathcal R}$ and $\mathcal G_\kappa$ act episode-wise on datasets:
\begin{equation}
\begin{aligned}
e_{\mathcal R}^{\kappa}
&=
\mathcal G_{\kappa}
\!\left(e_{\mathcal R}^{\mathrm{can}}\right)
=
\left(
\left\{
I_t^{\mathcal R},s_t^\kappa,a_t^\kappa
\right\}_{t=1}^{T},
\ell
\right),\\
\mathcal D_{\mathrm{KODA}}^\kappa
&=
\mathcal G_{\kappa}
\!\left(
\mathcal C_{\mathcal R}
\!\left(
\mathcal D_{\mathrm{human}}
\right)
\right),\quad
\mathcal D_{\mathrm{train}}^\kappa
=
\mathcal D_{\mathrm{robot}}^\kappa
\cup
\mathcal D_{\mathrm{KODA}}^\kappa .
\end{aligned}
\end{equation}
The mapping $\mathcal G_\kappa$ retains the canonical observations and instruction,
adding state features and optimized command labels in $\kappa$'s interface
as described in Sec.~\ref{sec:vla_training}. Converted and real-robot episodes share
a format and jointly train the VLA policy with parameters $\theta$ to predict
$\widehat a_t^\kappa
=
\pi_\theta(I_t^{\mathcal R},s_t^\kappa,\ell)$.

\subsection{Whole-Body Motion Generation}
\label{sec:kinematic_generation}

Given the camera poses and hand references in $e_h$,
Whole-Body Motion Generation outputs the simulator-checked canonical
whole-body motion $\{q_t^{\mathcal R}\}_{t=1}^{T}$.

\paragraph{Contact-Aware Hand Retargeting}
HAWOR~\citep{zhang2025hawor} estimates world-space MANO hand poses from
egocentric RGB. Following AnyTeleop~\citep{qin2023anyteleop}, KODA retargets
them to obtain initial robot wrist poses and active finger-joint values using MANO palm geometry and
a calibrated wrist-to-palm transform; distal mimic joints follow a fixed
coupling map~\citep{qin2023anyteleop}. Contact-aware refinement then adjusts
this initialization on reliable contact frames. To identify these frames, EfficientSAM3~
\citep{zeng2025efficientsam3} segments hand/forearm pixels, while nearby valid
non-hand depths provide object-surface samples. Reliable contact frames are
identified from multi-finger proximity, depth, and local surface consistency,
then temporally filtered; without reliable contact, the initialization is
retained.

At contact frames, wrist and finger poses are refined to preserve hand--object
geometry. Following OmniRetarget~\citep{yang2025omniretarget}, selected human
landmarks and corresponding robot-hand points are paired with the same nearby
object samples to form connected world-frame interaction meshes. Each vertex
descriptor is its offset from the uniformly weighted mean of its neighbors.
We optimize wrist poses and finger joints to minimize differences between
corresponding human and robot descriptors, preserving the demonstrated
hand--object geometry.
During optimization, object samples remain fixed, robot points are recomputed
by forward kinematics, wrist translations, local rotations, and finger
corrections are bounded, and deviations from initialization are penalized.
Corrections are interpolated between contact frames and held constant outside
their range. The optimized bilateral-wrist and finger trajectories are denoted
by $w_{1:T}^{\mathrm{ret}}$ and $f_{1:T}^{\mathrm{ret}}$, respectively; the
tracking stage uses $C_t^h$ and $w_{1:T}^{\mathrm{ret}}$ as sparse inputs.
See Appendix~\ref{app:motion_contact} for details.

\paragraph{Whole-body tracking via command optimization}

The source-camera poses are converted to torso poses using the
camera-to-torso transform from the target robot's URDF,
$p_t^{\mathrm{torso}}=\mathcal{T}_{\mathrm{cam}\to\mathrm{torso}}(C_t^h)$.
Combining $p_t^{\mathrm{torso}}$ with $w_t^{\mathrm{ret}}$, concatenating the
tracks in a fixed order, and applying a Savitzky--Golay filter
~\citep{schafer2011savitzky} yields $r_{1:T}^{\mathrm{ref}}$.

Kimodo~\citep{rempe2026kimodo} generates the initial whole-body configuration $q_{\mathrm{init}}^{\mathcal R}$
from the first reference frame. After kinematic and physics checks, this
configuration initializes the simulated rollout.

\begin{figure}[!t]
  \centering
  \includegraphics[width=\linewidth]{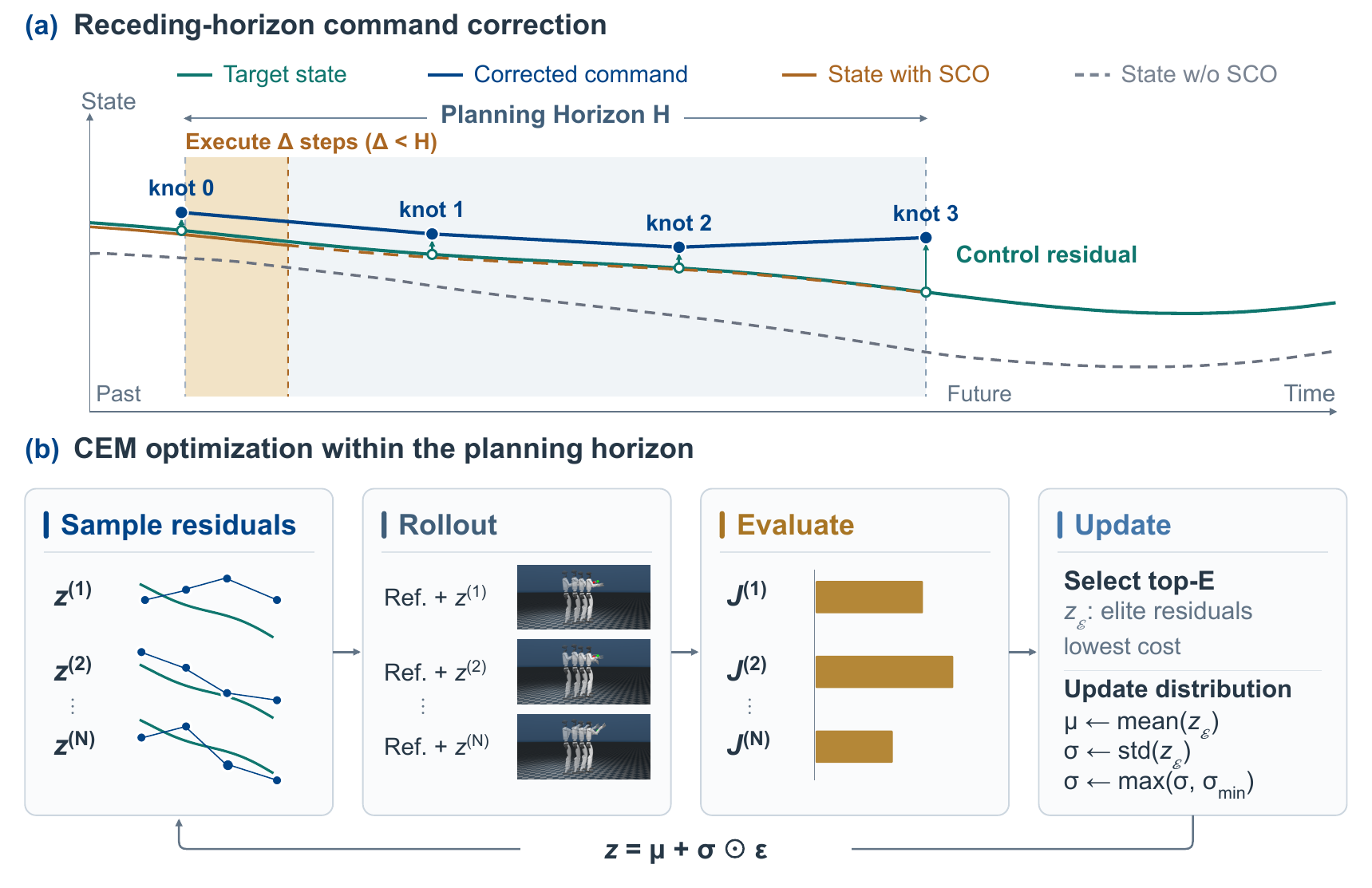}
  \caption{Sampling-based command optimization (SCO) uses CEM to optimize sparse pose residuals through sampling, rollout, evaluation, and proposal updates. (a) Receding-horizon correction with a planning horizon and executed prefix. (b) The CEM loop proceeds from sampling to rollout, evaluation, and distribution update; the feedback arrow returns to sampling.}
  \label{fig:cem-mpc}
\end{figure}

We add a three-point torso-and-bilateral-wrist tracking mode to ScaleBFM~
\citep{zeng2026scaling} and fine-tune the pretrained whole-body controller
$\phi_{\mathcal R}$ for the target embodiment. Its whole-body joint targets are
applied by PD control in MuJoCo/MJX, but direct tracking can accumulate
reference-tracking errors, as shown in TABLE~\ref{tab:motion_generation_error}. KODA therefore
uses the cross-entropy method (CEM)~\citep{pinneri2021sample} to optimize the sparse pose commands so that
the realized motion tracks $r_{1:T}^{\mathrm{ref}}$ more accurately while
remaining simulation-feasible. Candidate commands are evaluated by simulated
rollouts in a receding-horizon procedure detailed next and illustrated in
Fig.~\ref{fig:cem-mpc}; settings are reported
in Sec.~\ref{sec:real_robot_setup}.

At each replan, $\mathbf r^{\mathrm{ref}}$ spans $H$ steps from
simulated state $x_{\mathrm{cur}}$. The latent vector
$z\in\mathbb R^d$ contains 18 translation/rotation-vector coordinates
per knot for torso and bilateral-wrist SE(3) residuals.
CEM initializes diagonal-Gaussian parameters
$(\mu,\sigma)=(0,\sigma_0)$ and samples one residual for each of the
$N$ parallel MJX environments per iteration:
\begin{equation}
\begin{aligned}
z^{(n)}&=\mu+\sigma\odot\varepsilon^{(n)},
\qquad \varepsilon_i^{(n)}\overset{\mathrm{i.i.d.}}{\sim}\mathcal N(0,1),\\
\widetilde{\mathbf r}^{(n)}&=\operatorname{Decode}(\mathbf r^{\mathrm{ref}},z^{(n)}),
\end{aligned}
\label{eq:cem_sampling}
\end{equation}
where $\odot$ denotes element-wise multiplication, $i=1,\ldots,d$ indexes
latent coordinates, and $n=1,\ldots,N$ indexes candidates and environments.
The operator $\operatorname{Decode}$ applies the sampled residuals at the
reference knots, clips them to bounded physical ranges, and interpolates the
translation and rotation corrections between knots to form the complete
candidate pose-command trajectory $\widetilde{\mathbf r}^{(n)}$.
We roll out the $N$ candidates in parallel MJX environments:
\begin{equation}
\widehat{\mathbf q}^{(n)}=\operatorname{Rollout}_{\mathrm{MJX}}
(x_{\mathrm{cur}},\phi_{\mathcal R},\widetilde{\mathbf r}^{(n)}),\quad
\widehat{\mathbf r}^{(n)}=\Pi_{\mathcal B}(\widehat{\mathbf q}^{(n)}),
\label{eq:cem_rollout}
\end{equation}
Here $\widehat{\mathbf q}^{(n)}$ is the whole-body configuration trajectory
realized by the $n$-th candidate command in MJX, and
$\widehat{\mathbf r}^{(n)}=\Pi_{\mathcal B}(\widehat{\mathbf q}^{(n)})$ is the
corresponding torso-and-wrist pose trajectory. The set $\mathcal B$ contains
these torso and wrist bodies, and $\Pi_{\mathcal B}$ extracts their poses.
Each realized trajectory is scored against the unperturbed reference by
$J^{(n)}=d_{\mathcal B}(\widehat{\mathbf r}^{(n)},\mathbf r^{\mathrm{ref}})$,
a position/orientation tracking cost over these three links with
receding-horizon time weights.
Kinematic or physics violations are rejected. For the $E$ valid
lowest-cost samples indexed by $\mathcal E$, we update each dimension with
$\mu_i^+=\operatorname{mean}_{n\in\mathcal E}\!\left[z_i^{(n)}\right]$ and
$\sigma_i^+=\max\{\sigma_{\min,i},\operatorname{std}_{n\in\mathcal E}\!\left[z_i^{(n)}\right]\}$,
where the standard deviation uses divisor $E$ and is bounded below by
$\sigma_{\min,i}$.
After $I$ updates $(\mu,\sigma)\leftarrow(\mu^+,\sigma^+)$, we select
the lowest-cost valid rollout among the nominal, best sampled, and
final-mean candidates.
Only the first $\Delta$ steps are retained; the resulting simulator state
initializes the next replan, whose horizon advances by $\Delta$. Concatenated
prefixes form $q_{1:T}^{\mathcal R,\mathrm{SCO}}$ for constrained refinement.
See Appendix~\ref{app:motion_sco} for implementation details.

\paragraph{Constrained Arm Motion Refinement}
To reduce residual wrist-tracking error, temporal Mink IK~\citep{mink2024}
refines only the arm chain, including wrists. Contact-aware finger trajectories
are retained; the optimized floating base, torso, and lower body remain fixed
to preserve locomotion and balance motion. Refinement enforces joint limits,
per-step displacement and velocity bounds, and self-collision constraints,
while penalizing deviations from $q_{1:T}^{\mathcal R,\mathrm{SCO}}$.
Combining the refined arms and contact-aware fingers with the fixed non-arm
motion yields $q_{1:T}^{\mathcal R}$ for Robot Onboard View Generation.
Replay shows comparable controller trackability after refinement, as
quantified in Sec.~\ref{sec:motion_generation_action_instantiation}.
Further details are in Appendix~\ref{app:motion_refinement}.

\subsection{Robot Onboard View Generation}
\label{sec:robot_view_generation}

The optimized whole-body motion $q_{1:T}^{\mathcal R}$ from the previous
section determines the robot camera trajectory. Because camera poses map
camera coordinates to the world frame, the calibrated rigid torso--camera
transform gives
$C_t^{\mathcal R}=T_{\mathrm{cam}}(q_t^{\mathcal R})$.
KODA then synthesizes $I_t^{\mathcal R}$ from the source RGB frame $I_t^h$,
associated depth $D_t^h$, source-camera pose $C_t^h$, and target configuration
$q_t^{\mathcal R}$; all stages use this target camera pose and timestamp.

EfficientSAM3~\citep{zeng2025efficientsam3} predicts an arm/hand mask
$M_t^{\mathrm{arm}}$; depth inside it is invalidated to obtain
$\widetilde D_t^h$. KODA back-projects remaining valid pixels using
$\widetilde D_t^h$, calibrated camera intrinsics, and $C_t^h$, transforms them to the target camera
$C_t^{\mathcal R}$, and splats samples
with a metric z-buffer, producing $(I_t^{\mathrm{geo}},D_t^{\mathrm{geo}},
H_t^{\mathrm{geo}})$: the reprojected image, depth map, and hole mask.
MuJoCo renders $(I_t^{\mathrm{rob}},D_t^{\mathrm{rob}})$
from the same $q_t^{\mathcal R}$ and camera. At pixels where both depth maps
are valid, the nearer surface with smaller depth is retained: for pixel $u$,
$M_t^{\mathrm{rob}}(u)=1$ if and only if
$D_t^{\mathrm{rob}}(u)<D_t^{\mathrm{geo}}(u)$. If only one depth is valid, its
corresponding pixel is retained; neighboring valid depths are used only inside
geometric holes. The composite image is therefore
$I_t^{\mathrm{comp}}=M_t^{\mathrm{rob}}I_t^{\mathrm{rob}}+
(1-M_t^{\mathrm{rob}})I_t^{\mathrm{geo}}$.
DiffuEraser~\citep{li2025diffueraser} is applied only to residual holes
$H_t^{\mathrm{rem}}=H_t^{\mathrm{geo}}\setminus M_t^{\mathrm{rob}}$, namely,
pixels left uncovered by both the reprojected scene and the rendered robot.
Valid scene and robot pixels are frozen. The completed $I_t^{\mathcal R}$ is
indexed on the target control clock and paired with $q_t^{\mathcal R}$.
Together, the aligned observations $I_{1:T}^{\mathcal R}$, the whole-body
motion $q_{1:T}^{\mathcal R}$ generated above, and the instruction $\ell$ form
the canonical record $e_{\mathcal R}^{\mathrm{can}}$, which the next section
instantiates into controller-specific supervision.
See Appendix~\ref{app:observation} for details.

\begin{figure}[!t]
  \centering
  \includegraphics[width=0.9\textwidth]{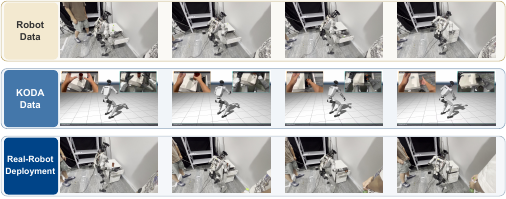}
  \caption{Qualitative example of the task ``Put the cup into the drawer.'' The rows show real-robot demonstrations (top), converted KODA data with aligned robot onboard observations (middle), and real-robot deployment of the learned VLA policy in the Robot-OOD setting (bottom). In the middle row, the main panels show the generated whole-body motion; the upper-left and upper-right insets show the source human view and the synthesized robot onboard view, respectively.}
  \label{fig:evaluation-overview}
\end{figure}
\subsection{Control Command Generation and VLA Training}
\label{sec:vla_training}

Given $e_{\mathcal R}^{\mathrm{can}}$, KODA generates controller-specific
command sequences for SONIC~\citep{luo2026sonic} and HEFT~\citep{liu2026heft}
on the Unitree G1 with BrainCo hands. Our command generation is related to test-time command
optimization~\citep{xie2026testtimecomputehumanoidbehavior},
which searches for controller-specific input commands through closed-loop
simulation to improve motion tracking. At 50\,Hz,
SCO optimizes a shared pelvis-pose and 29-body-joint residual space (hands
excluded) against fixed $q_{1:T}^{\mathcal R}$. The G1 reference contains
world-frame pelvis position and quaternion, 29 body-joint positions and
velocities, and world-frame link poses from forward kinematics. Each candidate
is decoded by the selected controller and rolled out in simulation with the
same cost for both controllers---joint-position MSE, 14-link pose error, reference deformation,
PD-target change, and terminal tracking penalties---and the same Control Command Generation settings
reported in Sec.~\ref{sec:real_robot_setup}.
The full search and objective are specified in
Appendices~\ref{app:interface_search} and~\ref{app:interface_cost}.

The resulting actions $a_t^\kappa$ use different controller interfaces. SONIC's
motion encoder maps the optimized motion references to 64-dimensional
motion tokens~\citep{luo2026sonic}. HEFT uses 29
absolute body-joint targets and a 9-dimensional pelvis code, giving 38
body-reference dimensions. The HEFT pelvis code represents roll $r$ and pitch
$p$ as $(\sin r,\cos r-1,\sin p,\cos p-1)$, together with a wrapped
next-step yaw change, three translation increments in the yaw-aligned frame,
and absolute height. Planar position and heading accumulate from an anchor,
whereas height is read directly. Both controllers append 12 hand-joint
targets, giving 76-dimensional SONIC and 50-dimensional HEFT actions. Pose
retargeting alone does not specify grasp force. KODA thresholds each
retargeted robot hand's mean finger curvature following
EgoHumanoid~\citep{shi2026egohumanoid}. For grasp, hand-joint targets are set to
a calibrated BrainCo grasp pose to encourage active closure; for release,
targets equal the current retargeted joint angles. The 12 targets are sent
separately to hand control.

Each converted episode records a 44-dimensional state vector $s_t^\kappa$ in a format shared by both controllers:
29 body-joint angles, three-dimensional rollout gravity rotated to the robot
base frame by the base quaternion, and 12 continuous BrainCo hand-joint
angles from contact-aware retargeting as geometric hand-state estimates.
Canonical images and the instruction are retained;
TABLE~\ref{tab:q3_tracking_error} reports wrist/camera pose residuals, and
TABLE~\ref{tab:human_data_scaling} reports co-training utility. GR00T~N1.7
predicts 40-step chunks of
$a_t^\kappa$ from $(I_t^{\mathcal R},s_t^\kappa,\ell)$ using converted and
real-robot episodes in the selected format. Changing the deployment interface
reruns control command generation and state recording while reusing the
canonical motion, observations, and instruction.
Interface and export details are in
Appendices~\ref{app:interface_body_labels}--\ref{app:interface_export}.

\section{Evaluation}

We evaluate whether KODA improves real-robot task success by augmenting fixed
robot datasets with human demonstrations, reusing the same canonical records
across SONIC and HEFT.
EgoHumanoid~\citep{shi2026egohumanoid} and HALOMI~\citep{zhao2026halomi}
use different control interfaces and do not directly provide compatible
whole-body state--action data. SONIC ablations assess the training benefits
of observation alignment and controller-specific supervision; tracking
diagnostics characterize motion fidelity and controller realization.

\begin{table}[!t]
\centering
\caption{Real-robot task success (\%) with increasing KODA data. Each controller uses 50 robot demonstrations per task. KODA demos denotes the number of converted human demonstrations per task; zero denotes robot-only training. Avg. is the macro-average over four tasks. Bold marks the best result within each controller, including ties.}
\label{tab:human_data_scaling}
\scriptsize
\setlength{\tabcolsep}{3.8pt}
\renewcommand{\arraystretch}{1.1}
\resizebox{\textwidth}{!}{%
\begin{tabular}{@{}ll*{10}{c}@{}}
\toprule
Controller & KODA demos
 & \multicolumn{2}{c}{Cup-to-basket}
 & \multicolumn{2}{c}{Parcel-to-shelf}
 & \multicolumn{2}{c}{Basket-to-shelf}
 & \multicolumn{2}{c}{Cup-to-drawer}
 & \multicolumn{2}{c}{Avg.} \\
\cmidrule(lr){3-4} \cmidrule(lr){5-6} \cmidrule(lr){7-8}
\cmidrule(lr){9-10} \cmidrule(lr){11-12}
& & Robot-ID & Robot-OOD
& Robot-ID & Robot-OOD
& Robot-ID & Robot-OOD
& Robot-ID & Robot-OOD
& Robot-ID & Robot-OOD \\
\midrule
\multirow{5}{*}{SONIC}
& $0$ & 55.00 & 0.00 & 90.00 & 15.00 & 55.00 & 0.00 & 50.00 & 0.00 & 62.50 & 3.75 \\
& $25$ & 55.00 & 10.00 & \textbf{100.00} & 25.00 & 50.00 & 15.00 & \textbf{70.00} & 15.00 & 68.75 & 16.25 \\
& $50$ & 65.00 & 20.00 & 90.00 & 65.00 & 60.00 & 10.00 & 65.00 & 0.00 & 70.00 & 23.75 \\
& $75$ & \textbf{70.00} & 35.00 & 90.00 & 70.00 & \textbf{65.00} & 30.00 & 60.00 & 30.00 & 71.25 & 41.25 \\
& $100$ & \textbf{70.00} & \textbf{40.00} & 90.00 & \textbf{75.00} & \textbf{65.00} & \textbf{45.00} & 65.00 & \textbf{35.00} & \textbf{72.50} & \textbf{48.75} \\
\midrule
\multirow{5}{*}{HEFT}
& $0$ & 65.00 & 0.00 & 85.00 & 20.00 & 65.00 & 0.00 & 60.00 & 0.00 & 68.75 & 5.00 \\
& $25$ & 65.00 & 5.00 & 90.00 & 35.00 & 45.00 & 20.00 & 65.00 & 20.00 & 66.25 & 20.00 \\
& $50$ & 60.00 & 30.00 & 80.00 & 55.00 & 65.00 & 25.00 & \textbf{70.00} & 15.00 & 68.75 & 31.25 \\
& $75$ & 70.00 & 40.00 & \textbf{95.00} & 65.00 & 60.00 & 35.00 & 65.00 & \textbf{40.00} & 72.50 & 45.00 \\
& $100$ & \textbf{80.00} & \textbf{50.00} & 90.00 & \textbf{75.00} & \textbf{70.00} & \textbf{40.00} & 65.00 & \textbf{40.00} & \textbf{76.25} & \textbf{51.25} \\
\bottomrule
\end{tabular}%
}
\end{table}

\subsection{Experimental Setup}
\label{sec:real_robot_setup}

\noindent\textit{Platform, data, and evaluation.}
We use a Unitree G1 with BrainCo dexterous hands on four whole-body tasks:
cup-to-basket, parcel-to-shelf, basket-to-shelf, and cup-to-drawer.
Each task has 50 real-robot demonstrations from the corresponding SONIC or
HEFT PICO teleoperation pipeline and 100 human egocentric demonstrations
captured with a chest-mounted iPhone using MobileEgo
Anywhere~\citep{palanisamy2026mobileego}.
All 400 human clips include iPhone-recorded depth. Both sources use 50\,Hz.
The scaling study evaluates SONIC~\citep{luo2026sonic} and
HEFT~\citep{liu2026heft}; component ablations use SONIC with 50 robot and
100 converted-human demonstrations per task.
Per controller configuration, GR00T N1.7~\citep{bjorck2025gr00t} trains on
all four tasks for 20,000 iterations with batch size 160. Within each task,
robot/human sampling is proportional to episode counts; other sampling
settings follow GR00T defaults.
We report one training run per configuration and
20 real-robot trials per task and condition.
This trial count matches EgoHumanoid's 20 trials per
setting~\citep{shi2026egohumanoid} and exceeds the 10 trials
per task and model reported in $\Psi_0$~\citep{wei2026psi0}.
Success requires completing the placement without a fall or
unrecovered manipulation failure.
Reported averages weight the four tasks equally.

\noindent\textit{Conditions covered by human demonstrations.}
Robot-ID uses object appearance and pose conditions present in the
robot demonstrations. Robot-OOD uses conditions absent from those
demonstrations: a green bottle is replaced by a brown bottle in
cup-to-basket and cup-to-drawer, a purple parcel by a white parcel in
parcel-to-shelf, and the basket handle is tilted by approximately
$30^\circ$ in basket-to-shelf.
The human training pool contains 80 shifted and 20 robot-matched clips
per task. Robot-OOD tests whether human demonstrations supply coverage
missing from the robot dataset, rather than zero-shot generalization
to conditions absent from all training data.
Data and evaluation details are in
Appendices~\ref{app:eval_data} and~\ref{app:eval_training}.

\noindent\textit{CEM settings.}
Both stages run at 50\,Hz and use CEM with 80 candidates per iteration,
8 iterations per replan, and a 40-step horizon.
Whole-Body Motion Generation uses 10 elites and executes 10 steps before replanning.
Control Command Generation with SONIC and HEFT uses 9 elites and executes
20 steps before replanning.

\subsection{Extending Robot Demonstration Coverage}
\label{sec:data_coverage}

Each task combines 50 fixed robot demonstrations with 0, 25, 50, 75, or 100
converted human demonstrations, randomly sampled independently at each budget
from its 100-clip pool without nesting. Policy architecture, training budget,
and evaluation protocol remain fixed.
TABLE~\ref{tab:human_data_scaling} shows that KODA provides its largest
gains on conditions missing from robot demonstrations.
With 100 converted demonstrations, SONIC Robot-OOD success rises from
$3.75\%$ to $48.75\%$ ($+45$ percentage points), and HEFT rises from
$5.00\%$ to $51.25\%$ ($+46.25$ points).
Robot-ID success also improves, from $62.5\%$ to $72.5\%$ for SONIC and
from $68.75\%$ to $76.25\%$ for HEFT.
Robot-OOD macro-average trends upward over tested budgets for both
controllers. Individual-task and Robot-ID results vary at intermediate
data sizes, so we focus on overall gains at the full data budget.
KODA thus supplies useful supervision for additional conditions covered
by human demonstrations without adding robot demonstrations.
Fig.~\ref{fig:evaluation-overview} illustrates converted data and
real-robot execution in the Robot-OOD setting for cup-to-drawer.
See Appendix~\ref{app:task_comparisons} for the other three tasks.
The same converted human demonstrations yield large Robot-OOD gains under
both SONIC and HEFT, demonstrating cross-controller reuse, as shown in
TABLE~\ref{tab:human_data_scaling}. The following ablations isolate
observation alignment and control command generation.

\subsection{Observation Alignment}
\label{sec:alignment_ablation}

We hold the converted demonstrations, state--action supervision, policy
architecture, training budget, and evaluation protocol fixed across four
tasks. No alignment retains the original human images. Fixed offset
uses a $0.10$\,m vertical camera shift with $\pm0.05$\,m training-time
perturbations, adapting the fixed-offset strategy
of~\citep{shi2026egohumanoid} to chest-mounted capture. Fixed offset and
Motion-based reprojection use only reprojection and hole inpainting;
Motion-based reprojection derives the target camera poses from the
canonical robot motion. Full alignment
additionally removes human arms and renders and composites the target robot.
See Appendix~\ref{app:eval_ablations} for details.

\begin{table}[!t]
\centering
\caption{Effect of observation alignment on SONIC policy success (\%). The four variants use the same demonstrations, state--action supervision, policy architecture, training budget, and evaluation protocol. Robot-ID and Robot-OOD are reported separately for each variant. Bold marks the highest success rate for each task and condition, including ties.}
\label{tab:view_alignment_ablation}
\small
\setlength{\tabcolsep}{2pt}
\renewcommand{\arraystretch}{1.2}
\begin{tabular*}{\textwidth}{@{\extracolsep{\fill}}l*{8}{c}@{}}
\toprule
Task
& \multicolumn{2}{c}{No alignment}
& \multicolumn{2}{c}{Fixed offset}
& \multicolumn{2}{c}{\shortstack{Motion-based\\reprojection}}
& \multicolumn{2}{c}{Full alignment} \\
\cmidrule(lr){2-3} \cmidrule(lr){4-5}
\cmidrule(lr){6-7} \cmidrule(lr){8-9}
& Robot-ID & Robot-OOD
& Robot-ID & Robot-OOD
& Robot-ID & Robot-OOD
& Robot-ID & Robot-OOD \\
\midrule
Cup-to-basket & 50 & 15 & 60 & 15 & 65 & 25 & \textbf{70} & \textbf{40} \\
Parcel-to-shelf & 85 & 35 & 80 & 45 & 85 & 65 & \textbf{90} & \textbf{75} \\
Basket-to-shelf & \textbf{65} & 20 & 60 & 25 & 60 & 35 & \textbf{65} & \textbf{45} \\
Cup-to-drawer & 50 & 5 & 55 & 10 & 55 & 25 & \textbf{65} & \textbf{35} \\
\midrule
Macro-average & 62.5 & 18.75 & 63.75 & 23.75 & 66.25 & 37.5 & \textbf{72.5} & \textbf{48.75} \\
\bottomrule
\end{tabular*}
\end{table}

Full alignment reaches $48.75\%$ Robot-OOD success versus $23.75\%$
with the adapted EgoHumanoid fixed-offset baseline~\citep{shi2026egohumanoid}
and $18.75\%$ without alignment (TABLE~\ref{tab:view_alignment_ablation}).
Motion-based reprojection reaches $37.5\%$.
Fixed offset, Motion-based reprojection, and Full alignment yield
Robot-ID point estimates of $63.75\%/66.25\%/72.5\%$, respectively.

\subsection{Controller-Consistent State--Action Supervision}
\label{sec:controller_instantiation_ablation}

The variant without SCO directly uses the generated canonical whole-body
motion as the tracking controller's reference, omitting optimization during
control command generation while retaining Whole-Body Motion Generation and
Robot Onboard View Generation.
Both variants use identical canonical trajectories, observations,
demonstrations, policy architecture, training budget, and evaluation
protocol. Since body states come from controller rollouts,
the comparison evaluates state--action supervision as a whole.
See Appendix~\ref{app:eval_ablations} for details.

\begin{table}[!t]
\centering
\caption{Effect of controller-specific command optimization on SONIC policy success (\%). Both variants use 50 robot and 100 converted-human demonstrations per task, with identical canonical trajectories and synthesized observations. Without SCO omits optimization only during Control Command Generation; With SCO uses optimized commands. Policy architecture, training budget, and evaluation protocol are held fixed. Bold marks the higher success rate for each task and condition.}
\label{tab:q3_policy_success}
\small
\setlength{\tabcolsep}{5pt}
\renewcommand{\arraystretch}{1.2}
\begin{tabular*}{\textwidth}{@{\extracolsep{\fill}}lcccc@{}}
\toprule
Task
& \multicolumn{2}{c}{Robot-ID}
& \multicolumn{2}{c}{Robot-OOD} \\
\cmidrule(lr){2-3}
\cmidrule(lr){4-5}
& Without SCO & With SCO
& Without SCO & With SCO \\
\midrule
Cup-to-basket
& 25 & \textbf{70}
& 0 & \textbf{40} \\
Parcel-to-shelf
& 70 & \textbf{90}
& 10 & \textbf{75} \\
Basket-to-shelf
& 50 & \textbf{65}
& 0 & \textbf{45} \\
Cup-to-drawer
& 50 & \textbf{65}
& 20 & \textbf{35} \\
\midrule
Macro-average
& 48.75 & \textbf{72.5}
& 7.5 & \textbf{48.75} \\
\bottomrule
\end{tabular*}
\par\smallskip
{\footnotesize\raggedright SCO: sampling-based command optimization. Macro-average weights the four tasks equally.\par}
\end{table}

TABLE~\ref{tab:q3_policy_success} shows that optimized control command generation
raises macro-average success from $48.75\%$ to $72.5\%$ Robot-ID and
from $7.5\%$ to $48.75\%$ Robot-OOD, gains of $23.75$ and $41.25$
percentage points. Both metrics improve on all four tasks.
Thus, controller-command generation also affects real-robot Robot-ID performance.
Without this optimization, Robot-ID success falls below the $62.5\%$
robot-only baseline in TABLE~\ref{tab:human_data_scaling}; the complete
pipeline reaches $72.5\%$. Controller-consistent supervision thus
matters to whether the added data benefits the policy. Together, the
ablations support aligning both observations and state--action
supervision with deployment.

\subsection{Motion Fidelity and Controller Trackability}
\label{sec:motion_generation_action_instantiation}

Motion-generation entries in TABLE~\ref{tab:tracking_error} average aggregate
errors from LAFAN~\citep{harvey2020robust} and 400 egocentric training clips.
LAFAN provides a check on highly dynamic motions; both controllers share evaluation data.

\begin{table}[!t]
\centering
\caption{Tracking errors for whole-body motion generation and controller rollouts. Position and orientation are reported separately in centimeters and radians; lower is better. Generation errors use the source motion as the reference, while rollout errors use the fixed canonical motion. Bold marks the smallest reported value for each target and error type, including ties.}
\label{tab:tracking_error}
\label{tab:motion_generation_error}
\label{tab:q3_tracking_error}
\small
\setlength{\tabcolsep}{3pt}
\renewcommand{\arraystretch}{1.2}
\begin{tabular*}{\textwidth}{@{\extracolsep{\fill}}ll*{6}{c}@{}}
\toprule
Stage & Target
& \multicolumn{2}{c}{Direct}
& \multicolumn{2}{c}{SCO}
& \multicolumn{2}{c}{SCO+IK} \\
\cmidrule(lr){3-4} \cmidrule(lr){5-6} \cmidrule(lr){7-8}
& & \shortstack{Position\\(cm)} & \shortstack{Orientation\\(rad)}
& \shortstack{Position\\(cm)} & \shortstack{Orientation\\(rad)}
& \shortstack{Position\\(cm)} & \shortstack{Orientation\\(rad)} \\
\midrule
\multirow{2}{*}{\shortstack[l]{Whole-body motion\\generation}}
& Wrist & 15.02 & 0.2046 & 3.03 & 0.1106 & \textbf{0.33} & \textbf{0.0296} \\
& Torso & 16.27 & 0.1302 & \textbf{5.39} & \textbf{0.0930} & \textbf{5.39} & \textbf{0.0930} \\
\midrule
\multirow{2}{*}{SONIC rollout}
& Wrist & 61.50 & 0.1777 & \textbf{2.88} & \textbf{0.1292} & -- & -- \\
& Camera & 60.90 & \textbf{0.0784} & \textbf{3.20} & 0.0808 & -- & -- \\
\midrule
\multirow{2}{*}{HEFT rollout}
& Wrist & 19.61 & 0.0677 & \textbf{2.94} & \textbf{0.0552} & -- & -- \\
& Camera & 19.51 & 0.0426 & \textbf{2.84} & \textbf{0.0337} & -- & -- \\
\bottomrule
\end{tabular*}
\par\smallskip
{\footnotesize\raggedright Direct uses unoptimized references. SCO uses CEM-optimized commands; SCO+IK adds Mink arm refinement during whole-body motion generation. A dash (--) denotes an unreported metric or variant.\par}
\end{table}

Wrist errors decrease to $3.03$\,cm/$0.1106$\,rad with SCO and
$0.33$\,cm/$0.0296$\,rad after Mink refinement.
Controller-specific SCO improves wrist/camera tracking except SONIC camera
orientation ($+0.0024$\,rad). Replaying the motions before and after Mink refinement with the same SONIC
tracker on both sources yields comparable errors: pelvis-relative non-arm/wrist
positions $2.60/1.59\!\to\!2.73/1.75$\,cm, torso orientation
$0.0588\!\to\!0.0579$\,rad, and non-arm joint RMSE
$0.0750\!\to\!0.0784$\,rad.
Metric and replay details are in Appendix~\ref{app:eval_metrics}.
TABLES~\ref{tab:view_alignment_ablation} and~\ref{tab:q3_policy_success}
show task benefits from observation alignment and controller-specific
supervision. TABLE~\ref{tab:human_data_scaling} supports the reusable records'
training utility across both controllers despite residual tracking mismatch.

\section{Limitations}
KODA does not yet fully reproduce fine-grained hand--object interactions,
limiting their fidelity and physical feasibility. In the future, we plan
to incorporate real-to-sim scene reconstruction and contact modeling to
better recover object geometry and interaction dynamics. We will also
broaden human data sources beyond egocentric recordings, including
third-person videos and motion capture, converting diverse human
demonstrations into canonical whole-body records for controller-specific
state--action generation and policy training.

\section{Conclusion}
KODA converts egocentric human demonstrations into canonical whole-body
motion and onboard observations, then generates controller-specific
state--action supervision for VLA training. Reusing these records with
fixed robot data improves VLA task success under SONIC and HEFT on four G1 tasks,
especially on conditions covered only by human demonstrations.
Reduced tracking errors and co-training gains despite residual mismatch
support expanding robot-policy coverage.

\clearpage
\bibliographystyle{plainnat}
\bibliography{B-migrated/reference}
\clearpage
\appendix
\numberwithin{equation}{section}
\numberwithin{table}{section}
\numberwithin{figure}{section}
\renewcommand{\theequation}{\Alph{section}.\arabic{equation}}
\renewcommand{\thetable}{\Alph{section}.\arabic{table}}
\renewcommand{\thefigure}{\Alph{section}.\arabic{figure}}
\section*{Appendix}
Appendix~\ref{app:motion} expands contact-aware retargeting, sparse command
optimization, and arm refinement. Appendix~\ref{app:observation} specifies
coordinate and visibility conventions. Appendix~\ref{app:interfaces} details
controller-specific search, action encoding, and state construction.
Appendix~\ref{app:evaluation} clarifies experimental conditions, comparisons,
and metric references. Appendix~\ref{app:qualitative} presents complete-clip
demonstration timelines. Appendix~\ref{app:task_comparisons} provides
Robot-ID, KODA, and Robot-OOD comparisons for the remaining three tasks.

\section{Whole-Body Motion Generation Details}
\label{app:motion}

This appendix expands Sec.~\ref{sec:kinematic_generation}. Source-camera
and bilateral hand references share a metric world frame. Retargeted wrist
and finger tracks $w_{1:T}^{\mathrm{ret}},f_{1:T}^{\mathrm{ret}}$ supply the
hand references; $r_{1:T}^{\mathrm{ref}}$ denotes the desired torso and wrist
poses. SCO produces realized whole-body motion
$q_{1:T}^{\mathcal R,\mathrm{SCO}}$. Arm refinement yields the canonical
floating-base, body, and hand sequence $q_{1:T}^{\mathcal R}$ while retaining
the hand references and non-arm motion.

\subsection{Contact-aware hand retargeting}
\label{app:motion_contact}

\paragraph{Initialization and contact evidence.}
HAWOR~\citep{zhang2025hawor} supplies world-space MANO hand estimates.
Retargeting following AnyTeleop~\citep{qin2023anyteleop} uses palm geometry and
a calibrated wrist-to-palm transform to initialize each robot wrist and its
active finger joints. Distal mimic joints are determined by the coupling map,
so they are not independent optimization variables. A correspondence between
selected human landmarks and robot-hand points provides the geometric basis
for refinement. The correspondences express analogous locations on the two
hands; they do not require their shapes or articulation ranges to coincide.

The hand/forearm segmentation removes human-hand pixels from the depth samples
used as object evidence. Nearby valid non-hand depths provide local object
surface points. Multi-finger proximity, depth validity, and local surface
consistency identify candidate contact frames, which are temporally filtered.
These checks determine where geometric refinement is supported by the
observation. When an entire sequence has no reliable contact, the initialized
hand trajectory is retained. Contact evidence is used to recover spatial hand--object
relationships; it is not a measurement of contact force.

\paragraph{Local wrist and finger corrections.}
At one contact frame, let $(p^0,R^0)$ be the initialized wrist pose and $f^0$
the active finger-joint vector for one hand. In the implemented six-active-joint
hand model, $\delta p,\delta\omega\in\mathbb R^3$ and
$\delta f\in\mathbb R^6$. We optimize
$\xi=(\delta p,\delta\omega,\delta f)\in\mathbb R^{12}$ using
\begin{equation}
\begin{aligned}
p(\xi)&=p^0+R^0\delta p,\\
R(\xi)&=R^0\operatorname{Exp}([\delta\omega]_\times),\\
f(\xi)&=\operatorname{clip}(f^0+\delta f,f_{\min},f_{\max}).
\end{aligned}
\label{eq:app_hand_delta}
\end{equation}
The initialized position $p^0$ is in the world frame, and $R^0$ maps wrist
coordinates to world coordinates. Both wrist corrections are expressed
relative to the initialized wrist frame.
Translations are in meters; rotation vectors and finger-joint angles are in
radians. The clip is componentwise, and $f_{\min},f_{\max}\in\mathbb R^6$
are the active-joint limits in the robot's MJCF model.
Here $[\delta\omega]_\times$ is the skew-symmetric matrix associated with the
rotation vector, and $\operatorname{Exp}$ maps it to a valid rotation. Forward
kinematics evaluates the robot landmarks under the corrected wrist and finger
configuration, including the fixed mimic-joint relationships. Thus wrist
placement and finger articulation affect the same geometric objective.

\paragraph{Shared interaction geometry.}
Let $P^h$ be the selected human landmarks, $P^{\mathcal R}(\xi)$ their robot
counterparts, and $O$ the object samples for this frame. Following the
interaction-mesh construction of OmniRetarget~\citep{yang2025omniretarget},
the ordered vertex arrays $V^h=[P^h;O]$ and
$V^{\mathcal R}(\xi)=[P^{\mathcal R}(\xi);O]$ share vertex correspondences
and connectivity. The implementation constructs adjacency from a Delaunay
tetrahedralization of the human landmarks and object samples, with a nearest-
neighbor fallback when tetrahedralization fails. Connectivity is then held
fixed throughout the frame's optimization and reused for the robot vertices.

Let $v_i\in\mathbb R^3$ be the world-frame position of the $i$th vertex of
$V$. For each vertex with a nonempty neighbor set $\mathcal N(i)$, its descriptor is
\begin{equation}
L_i(V)=v_i-\frac{1}{|\mathcal N(i)|}
\sum_{j\in\mathcal N(i)}v_j.
\label{eq:app_hand_descriptor}
\end{equation}
The uniform weights sum to one. Consequently, adding a common translation to
the vertices leaves their descriptors unchanged. Hand--object relationships
still constrain the wrist because the observed object vertices remain fixed
while the robot-hand vertices move. Object descriptors can also change when
their neighbors include hand vertices, even though the object positions
themselves are constant. The descriptor discrepancy therefore includes both
hand and object rows:
\begin{equation}
\xi^*=\arg\min_{\xi\in\mathcal A}
\lambda_L\sum_{i\in\mathcal V}
\bigl\|L_i(V^{\mathcal R}(\xi))-L_i(V^h)\bigr\|_2^2
+\mathcal L_{\mathrm{reg}}(\xi).
\label{eq:app_hand_objective}
\end{equation}
Here $\mathcal V$ indexes corresponding vertices and $\lambda_L$ weights
descriptor matching. Let $\xi^-=(\delta p^-,\delta\omega^-,\delta f^-)$ denote the
preceding solved contact-frame correction, with $\xi^-=0$ for the first
contact frame, and define $[a]_+=\max(a,0)$. The regularizer is
\begin{equation*}
\begin{aligned}
\mathcal L_{\mathrm{reg}}(\xi)={}&
\alpha_w^2\bigl(\|\delta p\|_2^2+\|\delta\omega\|_2^2\bigr)
+\alpha_f^2\|\delta f\|_2^2\\
&+\beta_w^2\bigl(\|\delta p-\delta p^-\|_2^2
+\|\delta\omega-\delta\omega^-\|_2^2\bigr)\\
&+\beta_f^2\|\delta f-\delta f^-\|_2^2\\
&+\gamma^2\bigl([\|\delta p\|_2-b_t]_+^2
+[\|\delta\omega\|_2-r_{\max}]_+^2\bigr).
\end{aligned}
\end{equation*}
The coefficients $\alpha_w,\alpha_f$ weight wrist and finger anchoring;
$\beta_w,\beta_f$ weight temporal changes in those corrections; and $\gamma$
weights the soft magnitude limits. These coefficients multiply residuals,
so their squares appear in the objective. The first two terms anchor the
correction to its initialization, the next two penalize changes from the
preceding contact solution, and the last term softly limits translation and
rotation magnitudes.
We use the following default weights:
\begin{center}
\small
\begin{tabular}{@{}lll@{}}
\toprule
Term & Parameter & Value \\
\midrule
Laplacian descriptor loss & $\lambda_L$ & $10$ \\
Wrist anchor residual & $\alpha_w$ & $0.10$ \\
Finger anchor residual & $\alpha_f$ & $0.018$ \\
Wrist temporal residual & $\beta_w$ & $12$ \\
Finger temporal residual & $\beta_f$ & $2$ \\
Soft-limit residual & $\gamma$ & $40$ \\
\bottomrule
\end{tabular}
\end{center}

The correction box at frame $t$ is
\begin{equation*}
\begin{aligned}
\mathcal A=\{\xi:\ &|\delta p_i|\le b_t,\quad
|\delta\omega_i|\le r_{\max}\quad(i=1,2,3),\\
&|\delta f_j|\le\delta f_{\max}\quad(j=1,\ldots,6)\}.
\end{aligned}
\end{equation*}
Here $b_t$ is the frame's translation bound in meters, $r_{\max}$ is the
rotation-vector coordinate bound in radians, and $\delta f_{\max}$ is the
finger-correction coordinate bound in radians. The default angular bounds are
$r_{\max}=\pi/15\ \mathrm{rad}$ ($12^\circ$) and
$\delta f_{\max}=0.28\ \mathrm{rad}$. These are componentwise box bounds;
the rotation-vector norm also enters the soft magnitude penalty above.
Active-joint limits are
applied separately by the clipping in Eq.~\eqref{eq:app_hand_delta}.
Let $d_{0,t}$ be the world-frame distance from the initialized palm center to
the centroid of the object samples used in the Laplacian graph, with
$d_{0,t}=0$ for an empty sample set. The default translation bound is
\begin{equation*}
b_t=
\begin{cases}
\min(d_{0,t},0.040\ \mathrm{m}),
& d_{0,t}\text{ is finite and }d_{0,t}\ge0.025\ \mathrm{m},\\
0.015\ \mathrm{m}, & \text{otherwise}.
\end{cases}
\end{equation*}
These coordinate bounds and soft magnitude penalties restrict displacement
when local depth evidence is incomplete.

\paragraph{Numerical refinement and temporal propagation.}
The contact solver stacks the descriptor, anchor, temporal, and soft-limit
residuals into a vector $\rho(\xi)$ so that $\|\rho(\xi)\|_2^2$ equals the
objective in Eq.~\eqref{eq:app_hand_objective}. In particular, descriptor
differences are multiplied by $\sqrt{\lambda_L}$; the remaining residual
blocks use $\alpha_w,\alpha_f,\beta_w,\beta_f,\gamma$ as defined above.
Let $J_\rho$ denote the Jacobian of this weighted residual vector with respect
to $\xi$. The fixed Laplacian matrix $L$ satisfies $(LV)_i=L_i(V)$, and
$I_3$ is the $3\times3$ identity. The unweighted descriptor Jacobian follows
by applying $L\otimes I_3$ to the robot-vertex Jacobian; object-vertex
derivatives are zero. Translation derivatives follow the wrist transform
analytically. Rotation and finger-joint derivatives use forward finite
differences with an increment of $10^{-4}$ rad.
Each iteration computes a damped linearized step,
\begin{equation}
\delta\xi=-(J_\rho^\top J_\rho+\lambda_d I)^{-1}
J_\rho^\top\rho,
\label{eq:app_hand_linearization}
\end{equation}
where $I$ is the $12\times12$ identity and $\lambda_d$ is the damping
coefficient, set to $10^{-4}$. The solver first
clips the step against the remaining coordinate bounds, then rescales it if
its Euclidean norm exceeds the default trust-region radius of $0.2$.
This norm uses the correction coordinates in meters and radians without
additional coordinate normalization. The corrected pose is re-evaluated
before the next iteration. The default iteration limits are $20$ for the
first contact frame and $10$ for subsequent contact frames. Iterations stop
earlier when successive squared-residual costs satisfy NumPy's default
\texttt{isclose} test or when the clipped and rescaled step has norm below
$2\times10^{-4}$; in the latter case, the small step is applied before
termination.
The preceding contact solution initializes the next contact-frame solve.

Within the contact optimizer, correction candidates are propagated to the
target clock by linear interpolation of translation and finger offsets
and shortest-path interpolation of the correction rotations on $SO(3)$
between contact intervals. Before the first
contact, the correction ramps from zero at the sequence start to the first
contact correction; after the last contact it ramps to zero at the sequence
end. These ramps apply when the corresponding non-contact frames exist.
With no contact frames, all corrections are zero. The propagated correction
coordinates are then smoothed with a separate Savitzky--Golay filter whose
default window contains nine frames and whose polynomial degree is two,
using polynomial edge evaluation (\texttt{interp} mode). Sequences shorter
than nine frames are left unchanged by this filter. Filtering does not impose
the prefilter endpoint values as equality constraints. This correction
filter is distinct from the sparse-reference filter described in
Sec.~\ref{app:motion_refinement}. The resulting
$f_{1:T}^{\mathrm{ret}}$ is retained as a continuous geometric hand reference;
the grasp and release command construction is a separate operation described
in Sec.~\ref{sec:vla_training}.

For the tracking pipeline, the wrist reference is assembled from the
candidate wrist poses over the interval between the first and last
contact frames. Outside this interval, the first-contact residual is
held before contact, and the last-contact residual is held after contact.
The held quantities are the world-frame translation offset and the local
rotation offset composed on the right of the nominal wrist orientation.
They are applied relative to the nominal wrist pose at each frame, so
holding these residuals does not fix the absolute wrist pose.
This boundary rule applies to the injected wrist reference and does not
replace the interpolated and smoothed finger trajectory. The assembled
three-point reference is subsequently filtered as described in
Sec.~\ref{app:motion_refinement}.

\subsection{Sparse pose-command optimization}
\label{app:motion_sco}

\paragraph{Search space and decoding.}
The source-camera poses are mapped to torso poses using the target robot's
URDF camera-to-torso transform. Together with the retargeted wrists, these
produce three ordered reference tracks: torso, left wrist, and right wrist.
We extend ScaleBFM~\citep{zeng2026scaling} with a three-point tracking mode
conditioned on the torso and bilateral wrist poses, and fine-tune the
pretrained controller for this mode using ScaleBFM's training data. The
resulting tracker receives these sparse pose commands and produces
whole-body joint targets for PD control. SCO changes the sparse commands
and evaluates their realized motion in simulation; it does not directly
search all robot joint configurations.

At each replanning instant, $K=10$ knots span a horizon of $H=40$ control
steps. For each tracked body $b$ and knot $k$, the latent contains translation
and rotation-vector residuals $(z^p_{kb},z^R_{kb})\in\mathbb R^6$.
The three bodies therefore contribute $18$ coordinates per knot and $180$
coordinates in total. Normalized coordinates are clipped to $[-1,1]$ before
physical scaling:
\begin{equation}
\delta p_{kb}=0.055\,\operatorname{clip}(z^p_{kb},-1,1),\qquad
\delta\omega_{kb}=0.0825\,\operatorname{clip}(z^R_{kb},-1,1).
\label{eq:app_sparse_scales}
\end{equation}
Translation is in meters and rotation is in radians. Both residuals are
expressed in the world frame. Let
$(p_{\tau b}^{\mathrm{ref}},R_{\tau b}^{\mathrm{ref}})$ denote the
unmodified reference pose of tracked body $b$ at horizon step
$\tau\in\{1,\ldots,H\}$, where $p_{\tau b}^{\mathrm{ref}}$ is its
world-frame position and $R_{\tau b}^{\mathrm{ref}}$ maps body-frame
vectors to the world frame.

The knots are uniformly spaced at
$\tau_k=1+(k-1)(H-1)/(K-1)$ for $k=1,\ldots,K$.
Let $A_{\tau k}$ be the piecewise-linear interpolation weight from knot
$k$ to horizon step $\tau$. For
$\tau_j\leq\tau\leq\tau_{j+1}$, the only potentially nonzero weights are
\[
A_{\tau j}=\frac{\tau_{j+1}-\tau}{\tau_{j+1}-\tau_j},
\qquad
A_{\tau,j+1}=\frac{\tau-\tau_j}{\tau_{j+1}-\tau_j}.
\]
The decoder interpolates the knot residuals and constructs the sparse
pose commands as
\begin{equation}
\begin{aligned}
\delta p_{\tau b}
&=\sum_{k=1}^{K}A_{\tau k}\,\delta p_{kb},
&
\delta\omega_{\tau b}
&=\sum_{k=1}^{K}A_{\tau k}\,\delta\omega_{kb},\\
p_{\tau b}^{\mathrm{cmd}}
&=p_{\tau b}^{\mathrm{ref}}+\delta p_{\tau b},
&
R_{\tau b}^{\mathrm{cmd}}
&=\operatorname{Exp}\!\left([\delta\omega_{\tau b}]_\times\right)
R_{\tau b}^{\mathrm{ref}}.
\end{aligned}
\label{eq:app_sparse_decode}
\end{equation}
Here $[\cdot]_\times$ is the skew-symmetric matrix operator satisfying
$[v]_\times x=v\times x$, and $\operatorname{Exp}$ is the matrix
exponential. The rotation-vector residuals are interpolated before
conversion to rotation matrices; left multiplication applies the resulting
rotation correction in the world frame.

The decoded poses
$(p_{\tau b}^{\mathrm{cmd}},R_{\tau b}^{\mathrm{cmd}})$ are supplied to
the three-point tracker. Their realized link poses are denoted by
$(\widehat p_{\tau b},\widehat R_{\tau b})$ in the tracking objective
below and are evaluated against the unchanged reference poses.
Joint references, pelvis guidance, and other body references are not
independently perturbed by this search. Their realized motion is
determined by the tracker and simulator.

\begin{table}[t]
\centering
\caption{SCO settings for Whole-Body Motion Generation. The residual and
proposal scales apply to this stage's three-point pose-command space.}
\label{app:motion_settings}
\begin{tabular}{@{}ll@{}}
\toprule
Quantity & Value \\
\midrule
Control frequency & $50$\,Hz \\
Planning horizon $H$ / executed prefix $\Delta$ & $40$ / $10$ steps \\
Temporal knots $K$ / latent dimension $d$ & $10$ / $180$ \\
Candidates $N$ / CEM updates $I$ / elites $E$ & $80$ / $8$ / $10$ \\
Initial standard deviation / floor & $1.6$ / $0.075$ \\
Translation / rotation scale & $0.055$\,m / $0.0825$\,rad \\
Position / orientation cost weight & $1.0$ / $0.1$ \\
Torso / left wrist / right wrist weight & $0.25$ / $1.0$ / $1.0$ \\
\bottomrule
\end{tabular}
\end{table}

\paragraph{Tracking objective.}
All candidates start from the same current simulator state and are scored
against the unmodified reference $\mathbf r^{\mathrm{ref}}$. Let
$(\widehat p_{\tau b},\widehat R_{\tau b})$ denote a candidate's realized
link pose at horizon step $\tau$, and let
$(p_{\tau b}^{\mathrm{ref}},R_{\tau b}^{\mathrm{ref}})$ be the corresponding
desired pose. The objective is
\begin{equation}
\begin{aligned}
J={}&\sum_{\tau=1}^{H}\omega_\tau
\sum_{b\in\mathcal B}\eta_b\Bigl[
\|\widehat p_{\tau b}-p_{\tau b}^{\mathrm{ref}}\|_2^2
+0.1\,\operatorname{angle}\!\left(
\widehat R_{\tau b}^{\top}R_{\tau b}^{\mathrm{ref}}\right)^2\Bigr],\\
\omega_\tau={}&\frac{H-\tau+1}{\sum_{j=1}^{H}(H-j+1)}.
\end{aligned}
\label{eq:app_sparse_cost}
\end{equation}
The body set $\mathcal B$ contains the torso and bilateral wrists.
$\operatorname{angle}$ gives the relative rotation angle in radians.
Link weights are $\eta_{\mathrm{torso}}=0.25$ and
$\eta_{\mathrm{left\ wrist}}=\eta_{\mathrm{right\ wrist}}=1$, without
normalization across links. The pelvis has zero weight; joint, velocity,
reference-deformation, and terminal penalties are inactive for this stage.
The decreasing time weights prioritize the prefix that will be executed next.
Scoring the realized motion against the unchanged target ensures that
changing the input command alone cannot be counted as improved tracking.

\paragraph{Diagonal CEM and simulator failures.}
CEM~\citep{pinneri2021sample} maintains a mean $\mu$ and a coordinate-wise
standard deviation $\sigma$, initialized to zero and $1.6$, respectively.
Each iteration samples $N$ latent vectors,
\begin{equation}
z^{(n)}=\mu+\sigma\odot\varepsilon^{(n)},\qquad
\varepsilon^{(n)}\sim\mathcal N(0,I_d),
\label{eq:app_cem_sample}
\end{equation}
then decodes and evaluates the corresponding commands in parallel rollouts.
For the $E$ lowest-cost candidates indexed by $\mathcal E$, the proposal
update is
\begin{equation}
\begin{aligned}
\mu_i^+&=\frac{1}{E}\sum_{n\in\mathcal E}z_i^{(n)},\\
\sigma_i^+&=\max\!\left\{0.075,
\sqrt{\frac{1}{E}\sum_{n\in\mathcal E}
\left(z_i^{(n)}-\mu_i^+\right)^2}\right\}.
\end{aligned}
\label{eq:app_cem_update}
\end{equation}
The divisor is $E$, giving the population standard deviation of the elite
coordinates. The proposal is diagonal, so the update does not estimate a
full covariance from a small elite set. Sampling operates in normalized
latent coordinates; physical clipping is applied by the decoder.

Non-finite simulator measurements or objective values are handled by
assigning a large finite failure cost before candidate ranking. CEM selects
the $E$ lowest-cost samples from all $N$ candidates without a separate
candidate-validity filter. The same failure-cost convention is used when
comparing the nominal command, the best sampled candidate, and the final
proposal mean. Tracking-error-based rejection is applied to complete
episodes after SCO, as described below.

\paragraph{Receding-horizon execution.}
One replanning update proceeds as follows:
\begin{enumerate}
\item Construct the next reference horizon and initialize the diagonal
proposal. All candidates share the current simulated starting state.
\item Sample, decode, and roll out $N$ sparse-command candidates. Score
them against the fixed target using the failure-cost convention above,
and update the proposal using the $E$ lowest-cost samples. Repeat for
$I$ updates.
\item Compare the nominal command, the best sampled candidate, and the
final proposal mean using their rollout costs, and retain the lowest-cost
choice.
\item Execute its first $\Delta=10$ steps, record the realized configurations,
and continue planning from the resulting simulator state.
\end{enumerate}
At $50$\,Hz, the horizon covers $0.80$\,s and each retained prefix covers
$0.20$\,s. The candidate count $N$ denotes sampled proposals per iteration;
nominal and final-mean evaluations are distinct comparison candidates.
Concatenating executed prefixes yields $q_{1:T}^{\mathcal R,\mathrm{SCO}}$.
Advancing from the realized state carries tracking deviations into the next
optimization instead of restarting each horizon at an ideal reference pose.

\paragraph{Episode acceptance before arm refinement.}
After SCO has generated the complete trajectory
$q_{1:T}^{\mathcal R,\mathrm{SCO}}$, we evaluate its mean wrist-position
tracking error before applying Mink arm refinement. Let
$\mathcal B_{\mathrm{wrist}}$ contain the left and right wrist links.
For each wrist $b$, let $p_{tb}^{\mathrm{SCO}}$ be its world-frame position
obtained by forward kinematics from $q_t^{\mathcal R,\mathrm{SCO}}$, and
let $p_{tb}^{\mathrm{ref}}$ be the corresponding unchanged wrist-position
target used during SCO. The episode-level error is
\begin{equation}
\bar e_{\mathrm{wrist}}
=
\frac{1}{2T}
\sum_{t=1}^{T}
\sum_{b\in\mathcal B_{\mathrm{wrist}}}
\left\|
p_{tb}^{\mathrm{SCO}}-p_{tb}^{\mathrm{ref}}
\right\|_2.
\label{eq:app_episode_wrist_error}
\end{equation}
This is the arithmetic mean of the Euclidean position errors over both
wrists and all frames of the complete episode. Episodes with
$\bar e_{\mathrm{wrist}}>0.075\,\mathrm{m}$ are discarded; retained
episodes proceed to Mink arm refinement. This acceptance criterion uses
wrist-position error only and is separate from the receding-horizon
optimization objective in Eq.~\ref{eq:app_sparse_cost}.

\subsection{Temporal smoothing and constrained arm refinement}
\label{app:motion_refinement}

\paragraph{Filtering on the rotation manifold.}
Savitzky--Golay filtering~\citep{schafer2011savitzky} smooths the three-point
reference positions in the common Z-up world frame, using a five-frame
window and a quadratic polynomial under the default settings. Orientation
filtering uses local tangent coordinates
so that smoothing does not produce invalid rotation matrices. For a window
around frame $t$, the neighboring rotations $R_j$ are represented relative
to $R_t$ as $\operatorname{Log}(R_t^\top R_j)^\vee$, using the principal
rotation vector with norm at most $\pi$. A local polynomial fit using the
actual timestamps is evaluated at the timestamp of frame $t$ to obtain
$\delta\omega_t^{\mathrm{SG}}$, and the filtered orientation is
\begin{equation}
R_t^{\mathrm{SG}}=R_t\operatorname{Exp}
\bigl([\delta\omega_t^{\mathrm{SG}}]_\times\bigr).
\label{eq:app_rotation_filter}
\end{equation}
The superscript $\vee$ converts a skew-symmetric matrix to its rotation
vector. Near a segment boundary, the full window shifts inside the segment
and the fitted polynomial is still evaluated at the target frame's timestamp.
The implementation splits the sequence where an adjacent timestamp gap
exceeds three times the median sampling interval of the full sequence;
segments shorter than five frames are left unchanged. Position
filtering uses the same actual-time polynomial fit and boundary handling.
Position and orientation filtering therefore use the same temporal reference
while respecting their different geometry. Kimodo~\citep{rempe2026kimodo}
provides an initial whole-body configuration from the first sparse reference
frame; the checked configuration initializes whole-body simulation.

\paragraph{Arm-only refinement and final assembly.}
SCO reduces sparse-reference error through the controller's dynamics, but
its realized wrists can still deviate from the retargeted manipulation
reference. Temporal Mink IK~\citep{mink2024} addresses this residual by
adjusting the arm chain, including the wrists. The optimized floating base,
torso, and lower-body trajectory are held fixed, and the contact-aware finger
trajectory is retained. Joint limits, per-step displacement and velocity
bounds, and self-collision constraints restrict the arm updates. A posture
penalty discourages unnecessary departures from the SCO configuration.
Temporal restrictions are applied across successive refined states so that
reducing a frame's wrist error does not permit an unconstrained joint jump.

The final configuration combines the refined arms, retargeted fingers, and
fixed non-arm motion. It supplies both the canonical motion reference and
the body trajectory used for robot-view rendering. Holding the non-arm
configuration fixed preserves those kinematic tracks; changes to the arms
can still affect the response of a whole-body controller. The replay
comparison in Sec.~\ref{sec:motion_generation_action_instantiation} therefore
evaluates trackability before and after refinement. These measurements assess
the demonstrated trajectories under the evaluated controller, without
assuming that inverse kinematics alone establishes dynamic feasibility for
arbitrary motions or deployment interfaces.

\section{Coordinate Conventions and Observation Synthesis}
\label{app:observation}

This appendix expands the geometric operations in
Sec.~\ref{sec:robot_view_generation}. The recovered human scene and the rendered
robot must be expressed in the same target-camera frame before their depths can
be compared. The target camera is obtained from the canonical whole-body motion,
so scene reprojection and robot rendering use one pose at each time index.

\subsection{Reference frames and the canonical record}
\label{app:observation_frames}

Let $T_{A\leftarrow B}\in\mathrm{SE}(3)$ map homogeneous coordinates in
frame $B$ into frame $A$. The world frame is denoted by $W$, the robot torso
frame by $B$, and its camera frame by $C$. A camera-to-world pose has the form
\begin{equation}
C_t=\begin{bmatrix}R_t&p_t\\0^\top&1\end{bmatrix},
\qquad
C_t^{-1}=\begin{bmatrix}R_t^\top&-R_t^\top p_t\\0^\top&1\end{bmatrix}.
\label{eq:app_view_pose}
\end{equation}
Here $R_t\in\mathrm{SO}(3)$ maps camera-frame vectors to the world frame,
and $p_t\in\mathbb R^3$ is the camera origin expressed in world coordinates.
Thus $C_t$ maps camera-frame points to world coordinates, whereas $C_t^{-1}$
maps world points into the camera frame for projection.
All depths and translations in the geometric stages use a consistent metric
scale. Image coordinates and calibrated intrinsics must refer to the same
image size. The control clock provides the common time index for source
observations, recovered geometry, canonical configurations, and the synthesized
observations stored with them.

Write $T_{B\leftarrow C}$ for the fixed camera-to-torso transform from the
target robot's calibration and URDF. We express the source and robot camera poses in a common world frame
and use a consistent optical-frame convention. To construct the torso
reference, we take the source camera pose $C_t^h$ as the desired robot
camera pose. The torso reference and the camera pose derived from canonical robot
motion are then given by
\begin{equation}
\begin{aligned}
T_{W\leftarrow B}^{\mathrm{ref}}(t)&=C_t^hT_{B\leftarrow C}^{-1},\\
C_t^{\mathcal R}&=T_{W\leftarrow B}(q_t^{\mathcal R})T_{B\leftarrow C}.
\end{aligned}
\label{eq:app_view_rigid_chain}
\end{equation}
Here $T_{W\leftarrow B}(q_t^{\mathcal R})$ is the torso-to-world transform
obtained by forward kinematics at the canonical robot configuration.
The first line expresses the torso target $p_t^{\mathrm{torso}}$ from
Sec.~\ref{sec:kinematic_generation} as a homogeneous transform; the second
uses the generated configuration to define the rendering camera. The resulting robot camera trajectory can differ from the source camera
trajectory because whole-body tracking may not exactly realize the torso
reference under the robot's kinematic and control constraints. Arm refinement
changes the rendered arm geometry while leaving the torso and camera
trajectory fixed. In particular, the observation stage uses the second line rather
than substituting the source camera path or a constant vertical displacement.

The canonical record retains $(I_t^{\mathcal R},q_t^{\mathcal R},\ell)$ for a
fixed target embodiment and camera convention. Its independence from the
deployment interface means that these quantities are retained when the
subsequent state--action mapping changes between SONIC and HEFT. A tracking
controller is still used to construct the canonical motion. The body state
recorded during deployment-controller replay can differ from the canonical
configuration; the interface appendix specifies this distinction in detail.

\subsection{Metric back-projection and target-camera visibility}
\label{app:observation_projection}

Let $K_h$ and $K_{\mathcal R}$ denote the source and target intrinsics. The following equations assume distortion-free pinhole image coordinates
and zero-skew intrinsics, with source RGB and depth registered to the same
pixel grid. A
source pixel is written as $u=(u_x,u_y)$, with homogeneous coordinate
$\bar u=(u_x,u_y,1)^\top$. EfficientSAM3~\citep{zeng2025efficientsam3}
provides the source arm/hand mask. Invalidating the masked depth before
projection prevents the source limbs from entering the reconstructed scene.
Only unmasked source pixels with finite, positive optical-axis depth are
back-projected. For each such sample,
\begin{equation}
\begin{aligned}
p_h(u)&=\widetilde D_t^h(u)K_h^{-1}\bar u,\\
\begin{bmatrix}p_{\mathcal R}(u)\\1\end{bmatrix}
&=(C_t^{\mathcal R})^{-1}C_t^h
\begin{bmatrix}p_h(u)\\1\end{bmatrix},\\
v_x(u)&=f_x^{\mathcal R}\frac{[p_{\mathcal R}(u)]_x}{[p_{\mathcal R}(u)]_z}
+c_x^{\mathcal R},\qquad
v_y(u)=f_y^{\mathcal R}\frac{[p_{\mathcal R}(u)]_y}{[p_{\mathcal R}(u)]_z}
+c_y^{\mathcal R}.
\end{aligned}
\label{eq:app_view_project}
\end{equation}
Here $\widetilde D_t^h$ is the source depth after invalidating the human
arm/hand mask. The points $p_h(u)$ and $p_{\mathcal R}(u)$ are expressed in
the source-camera and target-camera frames, respectively; $v(u)=(v_x(u),v_y(u))$
is the projected target-image location.
The quantities $f_x,f_y$ are focal lengths in pixels and $c_x,c_y$ are principal-point
coordinates. The source depth is a $z$ coordinate, rather than the Euclidean
distance from the camera center. After transformation, the target depth used
for visibility is $[p_{\mathcal R}(u)]_z$, rather than the original source
depth. Valid projected samples have finite coordinates and positive target
$z$ and contribute only inside the target image.

Forward splatting maps source colors into the target grid. Under the default settings, each projected location is rounded to
the nearest integer pixel center and splatted over its $3\times3$
neighborhood, clipped to the target image. Let
$\mathcal S_t(v)$ denote the valid source samples whose projected splats cover
target pixel $v$. The metric z-buffer selects the smallest target-camera depth
among the valid reprojected samples covering each target pixel,
\begin{equation}
D_t^{\mathrm{geo}}(v)=\min_{u\in\mathcal S_t(v)}[p_{\mathcal R}(u)]_z,
\qquad
V_t^{\mathrm{geo}}(v)=\mathbf 1[\mathcal S_t(v)\ne\emptyset].
\label{eq:app_view_zbuffer}
\end{equation}
Thus, $V_t^{\mathrm{geo}}$ indicates valid reprojected scene support;
visibility is resolved among the available source samples. The minimum is used only where $V_t^{\mathrm{geo}}=1$; elsewhere the depth is
invalid. Color is taken from a sample at the nearest surface, with numerical
ties resolved consistently. This visibility ordering prevents a farther
surface from overwriting a nearer one merely because it is processed later.
The geometric hole mask is $H_t^{\mathrm{geo}}=1-V_t^{\mathrm{geo}}$.
It includes areas exposed by source-arm removal and areas that were not
observed from the original viewpoint. Reprojection changes where visible
source evidence is placed; it does not supply the appearance of these newly
exposed regions.

\subsection{Robot rendering, compositing, and restricted completion}
\label{app:observation_composite}

MuJoCo renders robot color and depth from $q_t^{\mathcal R}$ with the same
$C_t^{\mathcal R}$ used in Eq.~\ref{eq:app_view_project}. The rendered depth
is represented in the same metric target-camera convention before visibility
comparison. Write $V_t^{\mathrm{rob}}$ for its binary validity mask. The
visible-robot mask is defined at every target pixel by
\begin{equation}
M_t^{\mathrm{rob}}(v)=
\begin{cases}
\mathbf 1[D_t^{\mathrm{rob}}(v)<D_t^{\mathrm{geo}}(v)],
  & (V_t^{\mathrm{rob}}(v),V_t^{\mathrm{geo}}(v))=(1,1),\\
1, & (V_t^{\mathrm{rob}}(v),V_t^{\mathrm{geo}}(v))=(1,0),\\
0, & (V_t^{\mathrm{rob}}(v),V_t^{\mathrm{geo}}(v))=(0,1),\\
0, & (V_t^{\mathrm{rob}}(v),V_t^{\mathrm{geo}}(v))=(0,0).
\end{cases}
\label{eq:app_view_order}
\end{equation}
If only one surface is valid, that surface supplies the pixel. When neither
is available, the pixel remains a hole. Neighboring valid scene depths may
supply local depth context inside geometric holes, as described in the main
pipeline; they do not replace the depth comparison at pixels with measured
scene support. The visible-robot mask, rather than the rendered silhouette
alone, controls the composite:
\begin{equation}
I_t^{\mathrm{comp}}=M_t^{\mathrm{rob}}I_t^{\mathrm{rob}}
+(1-M_t^{\mathrm{rob}})I_t^{\mathrm{geo}},
\qquad
H_t^{\mathrm{rem}}=H_t^{\mathrm{geo}}(1-M_t^{\mathrm{rob}}).
\label{eq:app_view_composite}
\end{equation}
Products here are pixel-wise, with binary masks broadcast over color channels.
The color arrays $I_t^{\mathrm{rob}}$ and $I_t^{\mathrm{geo}}$ use finite zero
placeholders wherever their respective validity masks are zero, so masked
arithmetic is defined even at holes. A branch-based selection of the valid
color gives the same composite. In particular,
$H_t^{\mathrm{rem}}=(1-V_t^{\mathrm{geo}})(1-V_t^{\mathrm{rob}})$:
neither source supplies a pixel precisely when both validity masks are zero.
A foreground scene object can occlude the robot, and a nearer robot surface
can occlude the reprojected scene. The residual-hole formula is equivalent
to the set difference in the main text when each binary mask is identified
with its support: only pixels uncovered by both sources remain to be completed.

DiffuEraser~\citep{li2025diffueraser} supplies appearance in
$H_t^{\mathrm{rem}}$. Denoting its completed prediction by $I_t^{\mathrm{fill}}$,
the final observation can be expressed as
\begin{equation}
I_t^{\mathcal R}=(1-H_t^{\mathrm{rem}})I_t^{\mathrm{comp}}
+H_t^{\mathrm{rem}}I_t^{\mathrm{fill}}.
\label{eq:app_view_completion}
\end{equation}
Thus, valid scene and rendered-robot pixels remain fixed. Completion has a
restricted appearance role after geometric visibility has been established;
it does not change the canonical configuration, target camera pose, or
action labels. The completed observation is stored with the same
$q_t^{\mathcal R}$ and instruction to form the canonical episode.

For the reported 400 egocentric training clips, source depth is recorded by
the iPhone. The RGB-only route in the framework instead uses metric depth
from Depth Anything 3~\citep{lin2025depthanything3} before the same geometric
operations. These are alternative sources for $D_t^h$, rather than two
separately evaluated training-data conditions. The observation ablation in
Sec.~\ref{sec:alignment_ablation} compares how views are aligned while keeping
the converted motion and state--action supervision fixed.

\section{Controller-Specific Command and State Interfaces}
\label{app:interfaces}

This appendix expands Sec.~\ref{sec:vla_training}. For each deployment
controller, KODA optimizes a body reference against the fixed canonical
trajectory, evaluates the realized motion, and records state--action
supervision. Table~\ref{app:interface_representations} distinguishes the
search representations from the quantities exported for VLA training.

\begin{table}[t]
\centering
\small
\caption{Representations used in control-command generation. Dimensions
refer to one knot or one time-indexed record, as indicated. The search
coordinates are shared, while body-action labels depend on the controller.}
\label{app:interface_representations}
\begin{tabular}{@{}p{0.22\linewidth}p{0.19\linewidth}p{0.53\linewidth}@{}}
\toprule
Quantity & Dimension & Contents and role \\
\midrule
Search residual & $35$ per knot & Pelvis translation and rotation-vector
corrections, plus 29 body-joint corrections. \\
Horizon search & $350$ & Ten residual knots interpolated over the horizon. \\
SONIC body label & $64$ & Motion token from the optimized motion reference. \\
HEFT body label & $38=29+9$ & Absolute body-joint targets and pelvis code. \\
Hand action & $12$ & Bilateral BrainCo targets, with grasp and release
handled separately. \\
Complete action & $76$ / $50$ & SONIC / HEFT body label with 12 hand targets. \\
Policy state & $44=29+3+12$ & Rollout body joints, base-frame gravity,
and retargeted hand angles. \\
\bottomrule
\end{tabular}
\end{table}

\subsection{Shared Body-Reference Search}
\label{app:interface_search}

Let $m_t^{\mathrm{can}}$ contain the canonical pelvis pose, the
29-dimensional body-joint vector $b_t^{\mathrm{can}}$, and the associated
link poses. The reference is represented on the 50\,Hz control clock;
joint velocities and link quantities are available with the motion record.
We use $\widetilde m_t$ for a candidate command reference and
$\widehat m_t^\kappa$ for the motion realized by controller $\kappa$.
These three quantities serve different roles: the canonical motion specifies
the desired behavior, the candidate is supplied through the controller's
reference interface, and the rollout determines its tracking cost.

For knot index $k=1,\ldots,K$, each temporal knot contains normalized residuals
\begin{equation}
z_k=\left(z_k^p,z_k^\omega,z_k^b\right)
\in\mathbb R^{3+3+29},\qquad
z=(z_1,\ldots,z_K),\quad K=10.
\label{eq:app_interface_search}
\end{equation}
Here $z_k^p,z_k^\omega\in\mathbb R^3$ and $z_k^b\in\mathbb R^{29}$
are dimensionless pelvis-translation, rotation-vector, and body-joint
corrections, respectively.
The resulting search vector $z\in\mathbb R^{350}$ is the variable
sampled by CEM for either controller. At each iteration, CEM draws
$N$ candidate vectors $z^{(n)}$, $n=1,\ldots,N$, from a diagonal
Gaussian distribution in this normalized knot space. Each sample
is deterministically decoded into physical corrections
$\delta_\tau^{(n)}$ at every control step through clipping, scaling,
and temporal interpolation in Eq.~\ref{eq:app_interface_interpolation}. Hands are
excluded from this body-reference search and retain the retargeted geometric
trajectory used for the hand-state and hand-command construction below.
Let $S_\kappa\in\mathbb R^{35\times35}$ denote the diagonal conversion from
normalized coordinates to physical translation, rotation, and joint
corrections. Both controllers use the same scales:
\begin{equation*}
S_{\mathrm{SONIC}}=S_{\mathrm{HEFT}}=
\operatorname{diag}\!\left(0.055 I_2,\;0.0275,\;0.0825 I_3,\;0.1375 I_{29}\right),
\end{equation*}
where $I_d$ is the $d$-dimensional identity matrix. The first three scale
entries are in meters; the rotation and joint entries are in radians.
For control-step index $\tau$ within the current horizon, let
$\delta_\tau=(\delta p_\tau,\delta\omega_\tau,\delta b_\tau)\in\mathbb R^{35}$
collect the dense physical corrections. With interpolation weights
$A_{\tau k}$, they are
\begin{equation}
\delta_\tau=\sum_{k=1}^{K}A_{\tau k}
S_\kappa\operatorname{clip}(z_k,-1,1),
\qquad \sum_k A_{\tau k}=1.
\label{eq:app_interface_interpolation}
\end{equation}
The knots are uniformly spaced over the horizon, and the nonnegative weights
$A_{\tau k}$ linearly interpolate neighboring knots. The clipping operation
acts elementwise. Let $p_\tau^{\mathrm{can}}$ and $R_\tau^{\mathrm{can}}$
be the canonical pelvis position and pelvis-to-world rotation. Translation
and rotation-vector corrections are expressed in the world frame and applied as
\begin{equation*}
\begin{aligned}
\widetilde p_\tau&=p_\tau^{\mathrm{can}}+\delta p_\tau,\\
\widetilde R_\tau&=\operatorname{Exp}([\delta\omega_\tau]_\times)
R_\tau^{\mathrm{can}},\\
\widetilde b_\tau&=b_\tau^{\mathrm{can}}+\delta b_\tau,
\end{aligned}
\end{equation*}
where $[v]_\times w=v\times w$ and $\operatorname{Exp}$ maps a
skew-symmetric matrix to a rotation. Body-joint corrections are followed by
elementwise clipping to joint limits when those limits are configured.
Forward kinematics then recomputes the link poses from the perturbed pelvis
and body joints, maintaining a single connected robot configuration.

The controller adapter encodes each candidate for simulation using SONIC
motion encoding or the HEFT representation below. CEM searches the common
physical reference space and evaluates each candidate through the selected
controller. Its 350 search coordinates differ from the 64-dimensional SONIC
token and 38-dimensional HEFT body label.

\subsection{Tracking Cost and Receding-Horizon Settings}
\label{app:interface_cost}

The shared objective evaluates 29 body joints and 14 links: the pelvis,
torso, and bilateral hip, knee, ankle, shoulder, elbow, and wrist links.
For two motion states $m,m'$, let $b,b'\in\mathbb R^{29}$ be their body-joint
angle vectors. Let $p_j,p'_j\in\mathbb R^3$ be the world-frame positions of
link $j$, and $R_j,R'_j\in SO(3)$ its link-to-world rotations, for
$j=1,\ldots,14$. Define
\begin{equation}
\begin{aligned}
e_q(m,m')&=\frac{1}{29}\|b-b'\|_2^2,\\
e_p(m,m')&=\frac{1}{14}\sum_{j=1}^{14}\|p_j-p'_j\|_2^2,\\
e_R(m,m')&=\frac{1}{14}\sum_{j=1}^{14}
\theta(R_j^\top R'_j)^2,\\
\theta(Q)&=\arccos\!\left(
\operatorname{clip}\!\left(\frac{\operatorname{tr}(Q)-1}{2},-1,1\right)
\right).
\end{aligned}
\label{eq:app_interface_errors}
\end{equation}
Here $Q\in SO(3)$ denotes a relative rotation and $\theta(Q)\in[0,\pi]$
its geodesic angle. Positions are in meters and angular quantities are in radians. Each link
contributes equally to the link averages; joint error is averaged over the
29 joint coordinates. The orientation term is the squared geodesic angle
of the relative rotation. Its quaternion implementation uses the absolute
quaternion inner product, so the two quaternion signs representing the same
rotation produce the same error.

For a horizon of $H$ steps, the per-scenario rollout cost is
\begin{equation}
\begin{aligned}
J_\kappa={}&\sum_{\tau=1}^{H}\Bigl[
e_q(\widehat m_\tau^\kappa,m_\tau^{\mathrm{can}})
+e_p(\widehat m_\tau^\kappa,m_\tau^{\mathrm{can}})
+0.5e_R(\widehat m_\tau^\kappa,m_\tau^{\mathrm{can}})\Bigr]\\
&+\mathcal L_{\mathrm{def}}
+0.01\sum_{\tau=2}^{H}
\|u_\tau^{\mathrm{PD}}-u_{\tau-1}^{\mathrm{PD}}\|_2^2
+\mathcal L_{\mathrm{terminal}},\\
\mathcal L_{\mathrm{def}}={}&\sum_{\tau=1}^{H}\Bigl[
0.1e_q(\widetilde m_\tau,m_\tau^{\mathrm{can}})
+0.1e_p(\widetilde m_\tau,m_\tau^{\mathrm{can}})
+0.05e_R(\widetilde m_\tau,m_\tau^{\mathrm{can}})\Bigr],\\
\mathcal L_{\mathrm{terminal}}={}&
5e_q(\widehat m_H^\kappa,m_H^{\mathrm{can}})
+5e_p(\widehat m_H^\kappa,m_H^{\mathrm{can}}).
\end{aligned}
\label{eq:app_interface_objective}
\end{equation}
The tracking terms compare realized motion with the fixed canonical target.
The deformation terms instead compare the candidate reference with that
target, penalizing unnecessary changes to the commands. The action-rate
term uses $u_\tau^{\mathrm{PD}}\in\mathbb R^{29}$, the low-level PD position
targets produced during rollout for the 29 body joints, in radians. Hand
targets are excluded. This term sums
squared changes across joints and has zero contribution at the first
horizon step. It does not measure changes between motion tokens or between
the normalized search vectors.

All running terms use uniform time weights of one. The horizon reduction is
a sum, while the joint and link reductions in
Eq.~\ref{eq:app_interface_errors} are averages. The terminal penalties add
weight to joint and link-position errors at the last step; no terminal
orientation term is included in this objective.

Control Command Generation uses $N=80$ candidates per CEM iteration,
$I=8$ iterations per replan, and $E=9$ elites. The planning horizon is
$H=40$ steps, or 0.80\,s, and the retained prefix is $\Delta=20$ steps,
or 0.40\,s. After executing that prefix in simulation, the next search
continues from its resulting state against the subsequent canonical
reference window. Candidate rollouts with non-finite joint, link, or action
values receive the failure cost $10^{12}$. The objective otherwise scores
the realized motion through the stated tracking terms; it does not add
a separate fall or contact heuristic. These settings apply to the second
optimization stage for both evaluated deployment controllers.

At the end of a sequence, the reference window is padded by repeating its
last frame to retain the full planning horizon. If $n_{\mathrm{rem}}$ control
steps remain, only $\min(\Delta,n_{\mathrm{rem}})$ steps are executed.

\paragraph{Dynamics randomization during command optimization.}
Following~\citep{xie2026testtimecomputehumanoidbehavior}, we evaluate
each command candidate in four randomized physics scenarios during
Control Command Generation.
At each replanning update, we sample dynamics multipliers and
initial-velocity offsets from Table~\ref{tab:command_generation_dr}
to construct the four scenarios.
The same scenario set is shared by all candidates and CEM iterations
within that update; a new set is sampled at the next update.
Each scenario rollout starts from the current execution state with
the sampled velocity offsets added.
The candidate reference and canonical target are shared across scenarios.

We evaluate Eq.~\ref{eq:app_interface_objective} separately in each
scenario, using its realized motion and PD targets.
For candidates with finite costs in all four scenarios, the CEM score
is the arithmetic mean of the four scenario costs.
The failure cost specified above is applied at the candidate level:
if any scenario produces a non-finite cost, the candidate score is
set to $10^{12}$.
The count $N=80$ denotes distinct command candidates, each evaluated
in all four scenarios.

After optimization, the selected prefix is executed under nominal
simulation dynamics. This execution provides the recorded body states
and the starting state for the next planning update.
SONIC and HEFT use the same randomization distributions, scenario
count, resampling rule, and cost aggregation.

\begin{table}[t]
\centering
\small
\setlength{\tabcolsep}{5pt}
\renewcommand{\arraystretch}{1.12}
\caption{Dynamics randomization for Control Command Generation.
Multipliers scale the corresponding nominal parameters and remain
fixed within each scenario rollout.
Velocity offsets are additive and sampled independently for each
component; $\mathcal{N}(0,\sigma^2)$ denotes a zero-mean Gaussian
distribution with standard deviation $\sigma$.}
\label{tab:command_generation_dr}
\begin{tabular*}{\linewidth}{@{\extracolsep{\fill}}lcc@{}}
\toprule
Parameter & Distribution & Unit \\
\midrule
Contact friction multiplier
& $\mathcal{U}(0.8,1.2)$ & -- \\
Joint armature multiplier
& $\mathcal{U}(0.8,1.2)$ & -- \\
Joint friction loss multiplier
& $\mathcal{U}(0.8,1.2)$ & -- \\
Actuation strength multiplier
& $\mathcal{U}(0.9,1.1)$ & -- \\
\midrule
Root linear velocity offset
& $\mathcal{N}(0,0.02^2)$ & $\mathrm{m/s}$ \\
Root angular velocity offset
& $\mathcal{N}(0,0.05^2)$ & $\mathrm{rad/s}$ \\
Joint velocity offset
& $\mathcal{N}(0,0.05^2)$ & $\mathrm{rad/s}$ \\
\bottomrule
\end{tabular*}
\end{table}

\subsection{SONIC and HEFT Body-Action Labels}
\label{app:interface_body_labels}

SONIC's motion encoder maps the optimized reference to a 64-dimensional
motion token~\citep{luo2026sonic}. This token is recorded as the body-action
label associated with the selected reference, whereas rollout body-joint
angles describe the resulting motion.

HEFT uses 29 absolute body-joint targets together with a nine-dimensional
pelvis code~\citep{liu2026heft}. Let $p_t\in\mathbb R^3$ and
$(\alpha_t,\beta_t,\psi_t)$ denote the optimized pelvis position and
roll, pitch, and yaw, with position expressed in the world frame. Let
$R_z(\psi)$ be the rotation about the world $z$ axis that maps vectors from
the yaw-aligned frame into the world frame. Define
\begin{equation}
\begin{aligned}
c_t={}&\bigl(\sin\alpha_t,\cos\alpha_t-1,
\sin\beta_t,\cos\beta_t-1,\delta\psi_t,
\delta p_t^\top,p_{t,z}\bigr)^\top,\\
\delta\psi_t={}&\operatorname{wrap}(\psi_{t+1}-\psi_t),\qquad
\delta p_t=R_z(\psi_t)^\top(p_{t+1}-p_t).
\end{aligned}
\label{eq:app_interface_heft_encode}
\end{equation}
Here $\operatorname{wrap}$ returns the principal signed angle
modulo $2\pi$. The row combines the current pose's roll, pitch, and absolute height with
the transition to the next pose. The translation increment is expressed
in the current yaw frame and contains three coordinates. A valid row
therefore needs a successor pelvis pose; this is part of the motion-code
definition rather than an extra body degree of freedom.
Encoding $L+1$ consecutive pelvis poses produces $L$ code rows; the final
input pose supplies the successor needed by the last row.

During decoding, planar position and heading accumulate from an anchor
$(\bar p_{0,xy},\bar\psi_0)$. Let $L$ be the number of code rows and use
local row indices $i=0,\ldots,L-1$, independent of the episode time index $t$.
With code components numbered from one, set
$\delta\psi_i=c_{i,5}$ and
$\delta p_i=(c_{i,6},c_{i,7},c_{i,8})^\top$. Then
\begin{equation}
\begin{aligned}
\bar\psi_i&=\bar\psi_0+\sum_{j=0}^{i-1}\delta\psi_j,\\
\bar p_{i,xy}&=\bar p_{0,xy}+
\sum_{j=0}^{i-1}\bigl[R_z(\bar\psi_j)\delta p_j\bigr]_{xy},
\qquad \bar p_{i,z}=c_{i,9},\\
\bar\alpha_i&=\operatorname{atan2}(c_{i,1},1+c_{i,2}),\qquad
\bar\beta_i=\operatorname{atan2}(c_{i,3},1+c_{i,4}).
\end{aligned}
\label{eq:app_interface_heft_decode}
\end{equation}
Empty sums are zero, and code components are numbered from one. Absolute
height directly defines the decoded vertical position; it is not obtained
by cumulatively adding the vertical increment. The final row's transition
leads beyond the returned sequence of current poses. This explains how
the code can contain both vertical displacement and absolute height
without integrating both into the same position. The decoded position $\bar p_i$ and roll--pitch--yaw angles
$(\bar\alpha_i,\bar\beta_i,\bar\psi_i)$ specify the pelvis pose
represented by code row $i$. Together with the corresponding
29 absolute body-joint targets, they recover the physical body
reference encoded by the HEFT body-action label.
HEFT uses pelvis-and-joint references together with the robot state
to generate tracking actions, which are mapped to body-joint PD
targets. The barred quantities describe reference poses, while
the realized motion is obtained from the controller rollout.

\subsection{Hand Targets and Recorded State}
\label{app:interface_hands_state}

Both body-action formats append 12 BrainCo hand-joint targets, six per
hand, yielding 76-dimensional SONIC actions and 50-dimensional HEFT actions.
The hand command is constructed from the retargeted robot hand geometry.
Following EgoHumanoid~\citep{shi2026egohumanoid}, the mean finger curvature
is thresholded separately for each hand to distinguish grasp from release.
For grasp, the command uses the calibrated BrainCo grasp pose to encourage
active closure. For release, it uses the current retargeted joint angles.
These targets are delivered separately to hand control. This construction
specifies the closure command without interpreting a retargeted geometric
pose as a measured grasp force.

The policy state has the same 44-dimensional format for both controllers:
\begin{equation}
s_t^\kappa=\left(
b_t^{\mathrm{roll},\kappa},\;
(R_t^{\mathrm{base},\kappa})^\top g^W,\;
f_t^{\mathrm{ret}}\right).
\label{eq:app_interface_state}
\end{equation}
Here $b_t^{\mathrm{roll},\kappa}\in\mathbb R^{29}$ contains rollout body-joint
angles and $f_t^{\mathrm{ret}}\in\mathbb R^{12}$ contains continuous retargeted
hand angles, both in radians. The rotation
$R_t^{\mathrm{base},\kappa}\in SO(3)$ maps base-frame vectors to the world
frame, and $g^W=(0,0,-1)^\top$ is the dimensionless unit gravity direction.
Thus $(R_t^{\mathrm{base},\kappa})^\top g^W$ gives the three-dimensional
projected-gravity observation in the rollout base frame. The state records the estimated hand geometry
even when a grasp action requests the calibrated closure target. Replacing
the state hand angles with that target would change the meaning of the
observation. Body-state values depend on the selected controller's rollout,
although the state format and retained hand estimates are shared.

\subsection{Episode Export and Observation Reuse}
\label{app:interface_export}

The procedure below summarizes the conversion from a canonical episode
to controller-specific supervision.

\begin{quote}
\small
\textbf{Controller-specific episode export}
\begin{enumerate}
\item Read the canonical motion, synthesized images, retargeted hand
trajectory, and instruction on their common temporal index.
\item For each receding-horizon window, decode CEM residuals into candidate
body references, encode them for controller $\kappa$, simulate, and evaluate
Eq.~\ref{eq:app_interface_objective} against the fixed canonical target.
\item Retain the selected prefix of up to 20 steps, shortened when fewer
steps remain, and continue from its resulting simulated state. Collect the
optimized reference and realized body motion.
\item Record the selected reference as SONIC tokens or HEFT body labels;
append the 12 grasp/release hand targets.
\item Assemble the 44-dimensional state from rollout body joints,
projected gravity, and retargeted hand angles.
\item Associate these states and actions with the retained canonical
images and instruction to obtain the controller-specific training episode.
\end{enumerate}
\end{quote}

Canonical-image reuse follows the fixed reference trajectory, while the
recorded body state follows the selected controller's realized motion.
Their correspondence is therefore approximate when tracking is imperfect.
The wrist and camera deviations in TABLE~\ref{tab:q3_tracking_error}
measure this residual mismatch for the evaluated controllers, and
TABLE~\ref{tab:human_data_scaling} evaluates the utility of the resulting
episodes for co-training. Changing from SONIC to HEFT repeats command
optimization, encoding, and state recording while retaining canonical
motion generation and view synthesis. The reuse rests on tracking the
shared reference sufficiently well for the evaluated setting; it does
not assert identical camera poses across controller rollouts.

\section{Experimental Protocol and Metric Definitions}
\label{app:evaluation}

This appendix specifies the data composition, controlled comparisons, and
interpretation of the evaluation metrics used in
Sec.~\ref{sec:real_robot_setup}. The task-success results remain in
TABLES~\ref{tab:human_data_scaling}--\ref{tab:q3_policy_success}; the motion
and rollout errors remain in TABLE~\ref{tab:tracking_error}.

\subsection{Demonstration Sources and Evaluation Conditions}
\label{app:eval_data}

The experimental platform is a Unitree G1 with BrainCo dexterous hands.
For each task and deployment controller, the robot dataset contains
50 demonstrations collected through the corresponding SONIC or HEFT PICO
teleoperation pipeline. The human pool contains 100 demonstrations per task,
recorded with a chest-mounted iPhone using MobileEgo
Anywhere~\citep{palanisamy2026mobileego}. All 400 human clips include
phone-recorded depth, and both demonstration sources use 50\,Hz sequences.
The same human canonical records supply the motion, synthesized observations,
and instructions for both controllers; command generation and state
recording produce the controller-specific training fields.

TABLE~\ref{app:eval_conditions} specifies the task goals and condition changes.
Robot-ID is defined relative to object appearance and pose conditions in the
robot demonstrations. Robot-OOD changes these conditions beyond that robot
dataset. Each task's human pool contains 80 shifted and 20 robot-matched
clips, so the shifted test conditions are represented in the human training
pool. This evaluates whether converted human data supplies coverage absent
from the robot demonstrations. It does not test conditions excluded from
both training sources. The 80:20 composition describes the complete human
pool; the independently sampled smaller budgets are not specified as having
that same proportion.

\begin{table}[!t]
\centering
\caption{Task goals and the changes defining Robot-OOD. Robot-ID uses the appearance and pose conditions represented in the robot demonstrations.}
\label{app:eval_conditions}
\small
\setlength{\tabcolsep}{5pt}
\renewcommand{\arraystretch}{1.18}
\begin{tabular*}{\textwidth}{@{\extracolsep{\fill}}p{0.18\textwidth}p{0.28\textwidth}p{0.46\textwidth}@{}}
\toprule
Task & Placement goal & Robot-ID to Robot-OOD change \\
\midrule
Cup-to-basket & Place the object in a basket & Green bottle replaced by a brown bottle \\
Parcel-to-shelf & Place the parcel on a shelf & Purple parcel replaced by a white parcel \\
Basket-to-shelf & Place the basket on a shelf & Basket handle tilted by approximately $30^\circ$ \\
Cup-to-drawer & Place the object in a drawer & Green bottle replaced by a brown bottle \\
\bottomrule
\end{tabular*}
\end{table}

\subsection{Training Budgets and Success-Rate Aggregation}
\label{app:eval_training}

For each controller configuration, GR00T N1.7~\citep{bjorck2025gr00t}
is trained jointly on all four tasks for 20,000 iterations with
batch size 160. Within each task, robot and converted-human sources
are sampled in proportion to their episode counts; other sampling
settings follow GR00T defaults.

The scaling study combines 50 fixed robot demonstrations with
$0$, $25$, $50$, $75$, or $100$ converted-human demonstrations per task.
Human subsets are sampled randomly and independently at each budget
without nesting. Policy architecture and training budget remain fixed.
All component ablations use SONIC with 50 robot and 100 converted-human
demonstrations per task.

Each configuration is trained once and evaluated in 20 real-robot
trials per task under each of the Robot-ID and Robot-OOD conditions.
A trial succeeds when the placement is completed without a fall or
unrecovered manipulation failure. We report the percentage of
successful trials for each task and the unweighted mean across
the four tasks, separately for each condition.

\subsection{Controlled Component Comparisons}
\label{app:eval_ablations}

TABLE~\ref{app:eval_variants} separates the observation variants from the
control-command variants. In the observation study, the demonstration set,
state--action supervision, policy architecture, training budget, and
real-robot evaluation protocol are fixed. Only the human-derived observation
construction changes. Fixed offset adapts the view-alignment strategy
of EgoHumanoid~\citep{shi2026egohumanoid} to chest-mounted capture, using a
$0.10$\,m vertical shift and $\pm0.05$\,m training-time perturbations.
Both reprojection-only variants include hole inpainting.

\begin{table}[!htbp]
\centering
\caption{Definitions of the component variants. All comparisons use
SONIC and the full human-data budget. The table describes changed
components; quantitative policy results are in
TABLES~\ref{tab:view_alignment_ablation}
and~\ref{tab:q3_policy_success}.}
\label{app:eval_variants}
\small
\setlength{\tabcolsep}{5pt}
\renewcommand{\arraystretch}{1.16}
\begin{tabular*}{\textwidth}{
    @{\extracolsep{\fill}}
    p{0.24\textwidth}
    p{0.70\textwidth}
    @{}
}
\toprule
Variant & Observation or command construction \\
\midrule

\multicolumn{2}{@{}l@{}}{\textbf{Observation alignment}} \\
\addlinespace[2pt]

No alignment
& Original human images. \\

Fixed offset
& Scene reprojection with a fixed vertical camera shift,
training-time perturbations, and hole inpainting. \\

Motion-based reprojection
& Scene reprojection and hole inpainting using camera poses
derived from the canonical robot motion. \\

Full alignment
& Motion-derived onboard views, additionally removing human arms
and rendering and compositing the target robot. \\

\midrule
\multicolumn{2}{@{}l@{}}{\textbf{Control Command Generation}} \\
\addlinespace[2pt]

Without SCO
& Canonical whole-body motion used directly as the tracking
controller's reference; command optimization omitted only
at this stage. \\

With SCO
& Controller-specific commands optimized to track the same
canonical motion. \\

\bottomrule
\end{tabular*}
\end{table}

Full alignment changes human-arm removal and robot rendering/compositing
together relative to motion-based reprojection. Consequently, this comparison
does not separately identify the effect of each image-processing operation.
The main interpretation concerns the complete alignment pipeline relative
to original human observations and the adapted fixed-offset baseline. The
intermediate variant describes a partial pipeline under the same training
setup.

In the command-generation study, both variants retain Whole-Body Motion
Generation and Robot Onboard View Generation. They use identical canonical
trajectories, observations, demonstrations, and policy-training settings.
The optimized commands are evaluated through the deployment controller,
and body-state features are recorded from the resulting rollouts. Changing
command generation can therefore change both action labels and body-state
features. This is a comparison of state--action supervision as a whole,
not an action-only intervention with all state inputs held fixed. These
component comparisons use SONIC; the data-scaling study evaluates reuse of
the complete converted records under both SONIC and HEFT.

\subsection{Motion References, Units, and Refinement Replay}
\label{app:eval_metrics}

The blocks of TABLE~\ref{tab:tracking_error} use different references.
Whole-body motion-generation errors compare the generated robot motion
with the source-derived motion references. Controller-rollout errors
compare the realized motion with the fixed canonical trajectory.
The former measures source-reference fidelity, while the latter measures
how closely a particular controller realizes the canonical motion.
Direct denotes references without command optimization at the evaluated
stage, SCO denotes CEM-optimized commands, and SCO+IK adds Mink arm
refinement during whole-body motion generation.

Position errors are reported in centimeters, and orientation errors and
joint-angle RMSE are reported in radians. The motion-generation block reports wrist and torso position
and orientation errors. The controller-rollout blocks report
wrist and camera position and orientation. The smallest reported values
are identified separately for each target and error type.
Improved position tracking does not necessarily imply improved
orientation tracking: SONIC's camera-orientation error is
$0.0784$\,rad for Direct and $0.0808$\,rad for SCO.
Dashes denote unreported metrics or variants.

Motion-generation entries average aggregate errors from
LAFAN~\citep{harvey2020robust} and the 400 egocentric training clips.
LAFAN provides an auxiliary check on highly dynamic motion, while the
egocentric clips connect the diagnostics to the converted policy-training
data. Both controllers share evaluation data. The reported aggregate
entries do not provide separate results for the two data sources.

TABLE~\ref{app:eval_mink_replay} expands the before/after-Mink replay
diagnostics described in
Sec.~\ref{sec:motion_generation_action_instantiation}.
The same SONIC tracker replays the pre-refinement and refined
trajectories on both data sources. For each replay, tracking errors
are measured against the corresponding input reference trajectory.
The diagnostics include pelvis-relative non-arm and wrist-position
errors, torso-orientation error, and non-arm joint-angle RMSE.
These replay errors measure controller trackability and are distinct
from motion-generation errors measured against the source-derived
references. Although refinement preserves the non-arm reference
trajectory, changes to the arm reference can affect the realized
whole-body motion and hence the non-arm replay errors.
The small changes indicate comparable SONIC trackability on the
evaluated trajectories; they do not constitute a downstream
policy-success ablation of Mink refinement.

\begin{table}[!htbp]
\centering
\caption{SONIC replay before and after Mink arm refinement on LAFAN
and the 400 egocentric clips. Each replay is evaluated against its
corresponding input reference trajectory.}
\label{app:eval_mink_replay}
\small
\setlength{\tabcolsep}{5pt}
\renewcommand{\arraystretch}{1.18}
\begin{tabular*}{\textwidth}{@{\extracolsep{\fill}}lccc@{}}
\toprule
Replay metric & Unit & Before refinement & After refinement \\
\midrule
Pelvis-relative non-arm position error
& cm & 2.60 & 2.73 \\
Pelvis-relative wrist-position error
& cm & 1.59 & 1.75 \\
Torso-orientation error
& rad & 0.0588 & 0.0579 \\
Non-arm joint RMSE
& rad & 0.0750 & 0.0784 \\
\bottomrule
\end{tabular*}
\end{table}

\clearpage
\section{Qualitative Demonstration Timelines}
\label{app:qualitative}

Each timeline spans one complete source clip through eight selected keyframes,
including its first and last frames. Read left to right, then top to bottom;
timestamps refer to the original 50\,Hz overlay video. Each frame retains the
human view (upper left), synthesized robot view (upper right), and offline
MuJoCo kinematic replay (main panel). These illustrate converted demonstrations;
they are not additional real-robot evaluation trials.

\begin{figure}[!ht]
  \centering
  \includegraphics[width=0.84\linewidth]{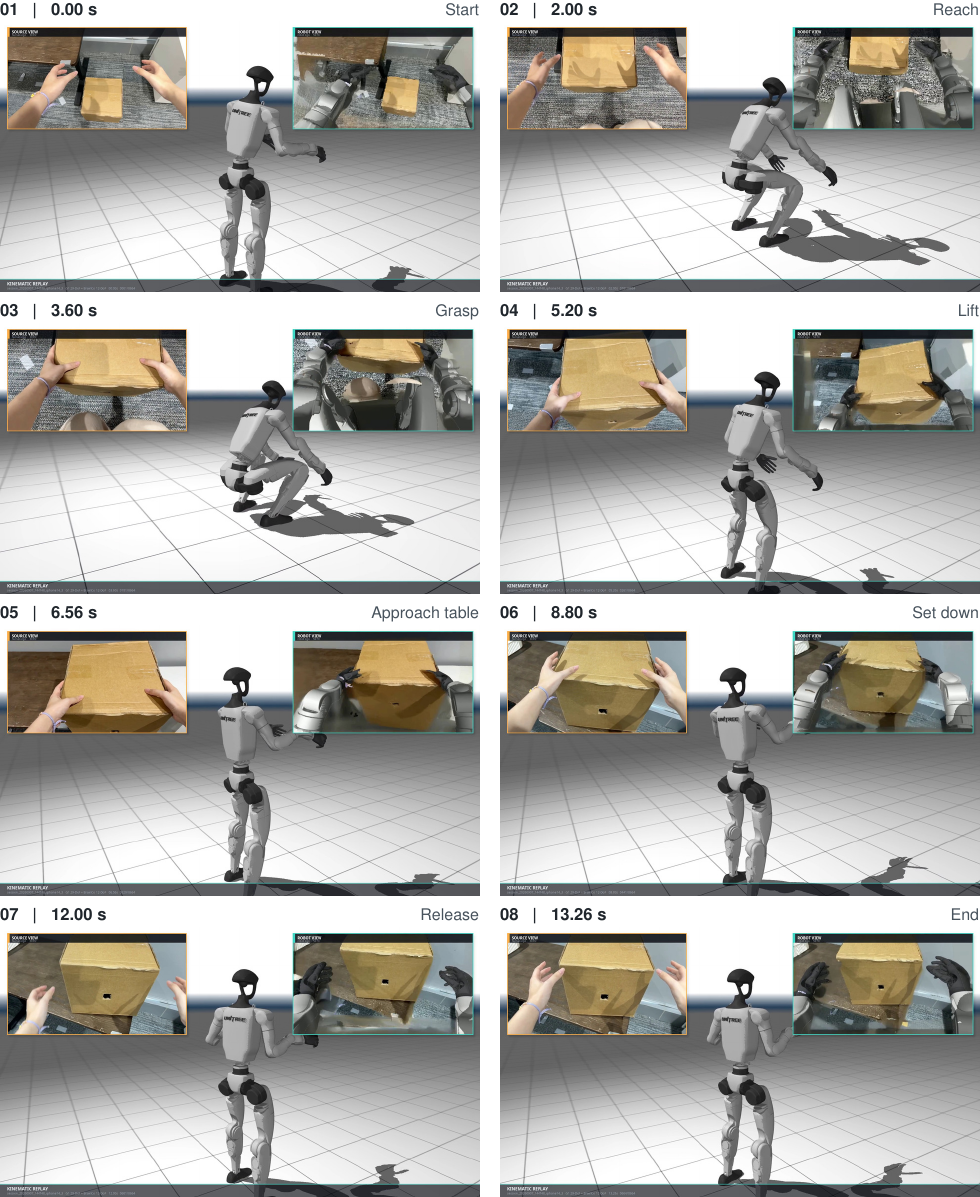}
  \caption{\textbf{Complete-clip timeline: bimanual box transfer.}
  Selected moments from a 13.28\,s clip show reaching toward a box, lifting
  and carrying it, and setting it on a table before withdrawing the hands.
  Frames 1--8 preserve chronological order and the paired human, robot-view,
  and whole-body-motion displays.}
  \label{fig:demo_timeline_box}
\end{figure}

\clearpage
\begin{figure}[!ht]
  \centering
  \includegraphics[width=\linewidth]{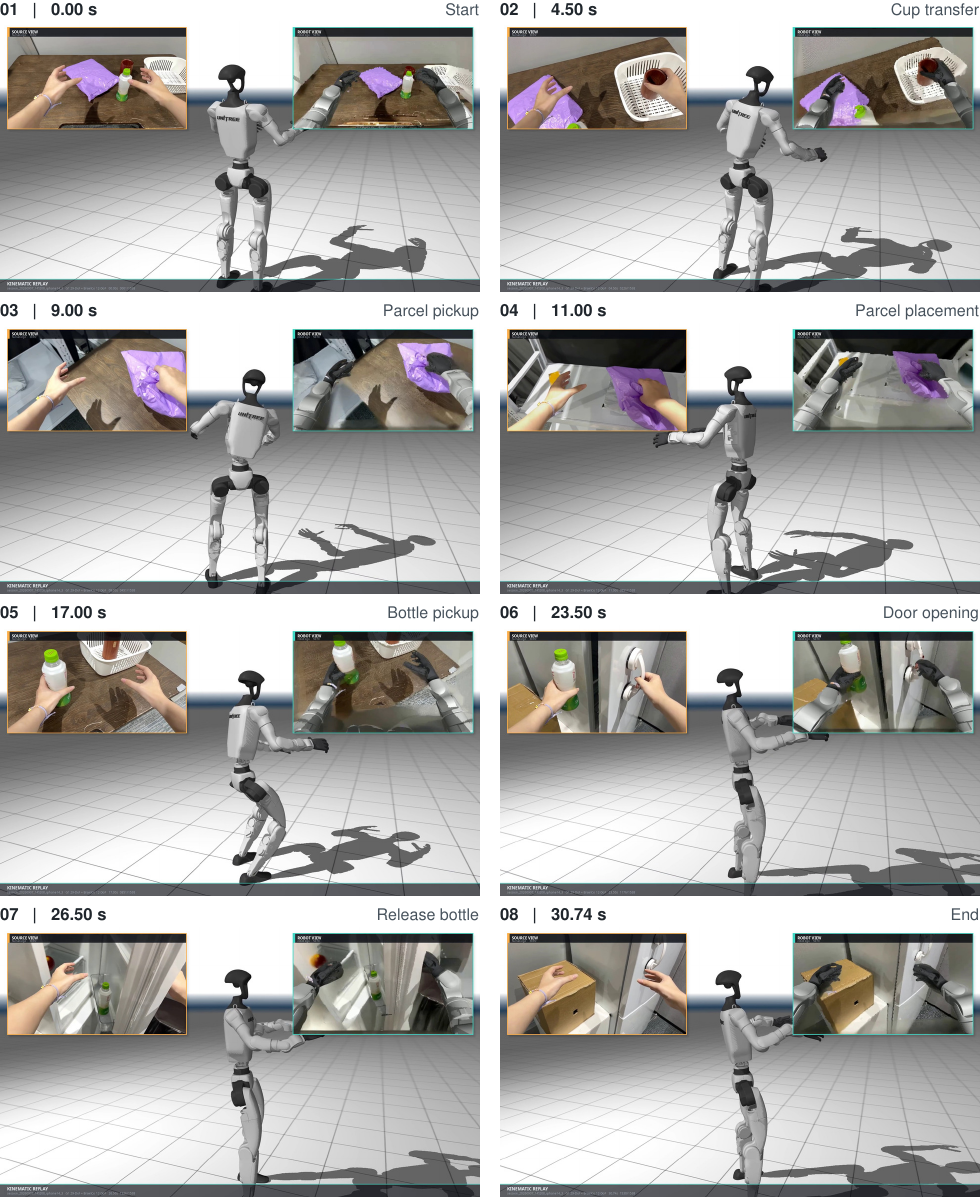}
  \caption{\textbf{Complete-clip timeline: sequential object manipulation.}
  Eight selected moments span a 30.76\,s clip: a cup is moved toward a basket,
  a parcel is transferred to a shelf, and a bottle is carried to a refrigerator
  and placed inside before the door is closed. The first and last video frames
  are included. Time advances left to right and then top to bottom; the three
  synchronized displays have the same meanings as in
  Fig.~\ref{fig:demo_timeline_box}.}
  \label{fig:demo_timeline_sequence}
\end{figure}

\clearpage
\section{Additional Task Demonstrations and Deployments}
\label{app:task_comparisons}

\begin{figure}[!ht]
  \centering
  \includegraphics[width=\linewidth]{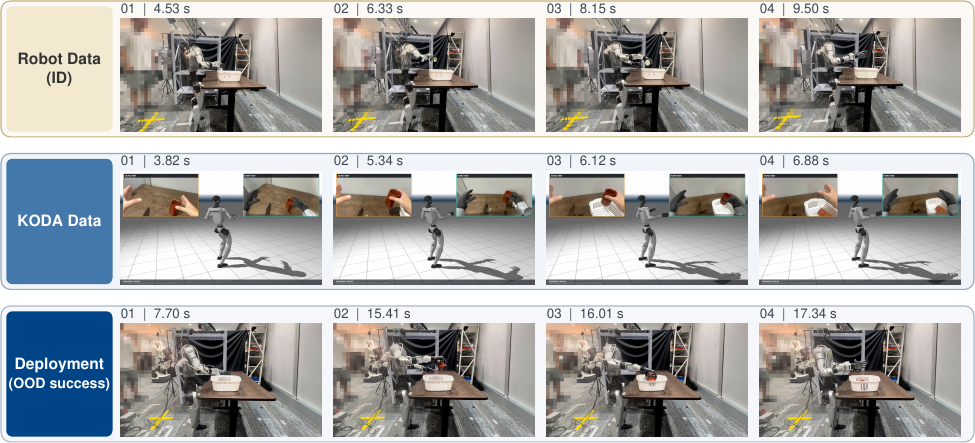}
  \caption{\textbf{Cup-to-basket.} Robot Data shows the green bottle used in
  Robot-ID. KODA Data covers the brown-bottle condition, and the deployment
  row shows a successful Robot-OOD placement into the basket. Each row
  progresses from grasping through transfer to placement.}
  \label{fig:task_cup_basket}
\end{figure}

\begin{figure}[!ht]
  \centering
  \includegraphics[width=\linewidth]{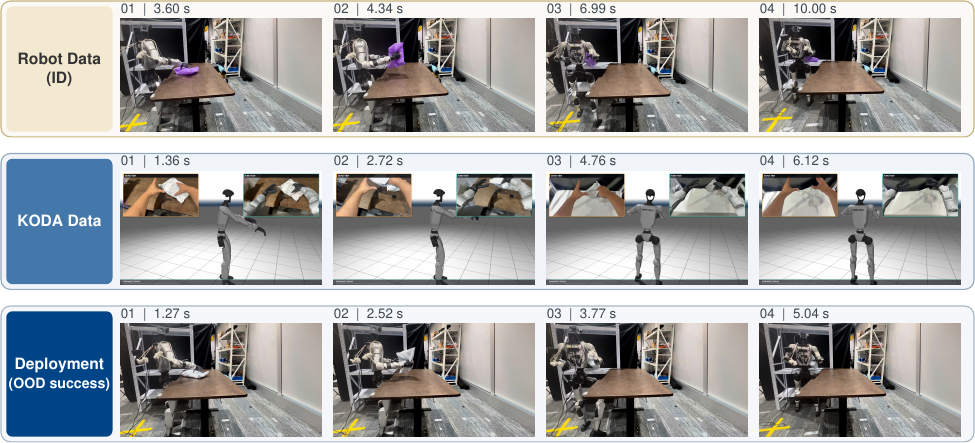}
  \caption{\textbf{Parcel-to-shelf.} The purple parcel in Robot Data is replaced
  by a white parcel in the KODA and Robot-OOD deployment examples. The
  keyframes show reaching, lifting and transferring the parcel, followed by
  placement on the shelf.}
  \label{fig:task_parcel_shelf}
\end{figure}

\begin{figure}[!ht]
  \centering
  \includegraphics[width=\linewidth]{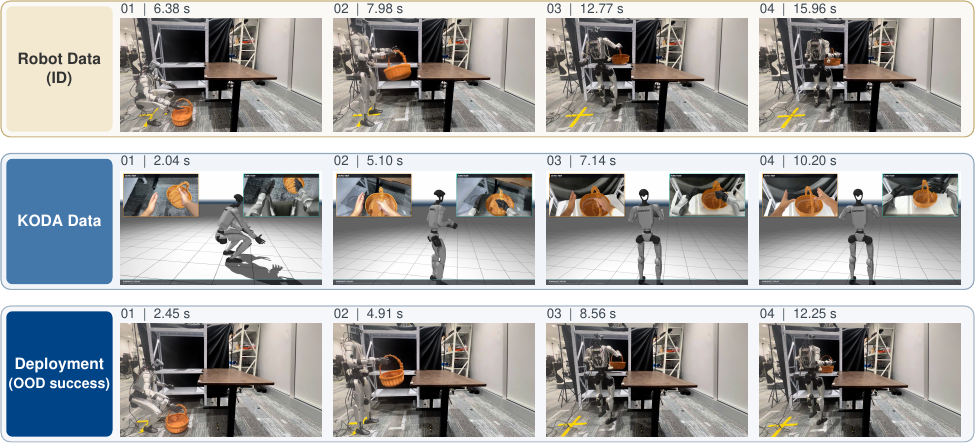}
  \caption{\textbf{Basket-to-shelf.} The Robot-OOD change is an approximately
  $30^\circ$ tilt of the basket handle relative to the Robot-ID condition.
  KODA Data supplies a corresponding converted demonstration; the deployment
  row shows grasping the handle, lifting the basket, and placing it on the
  shelf.}
  \label{fig:task_basket_shelf}
\end{figure}

Figures~\ref{fig:task_cup_basket}--\ref{fig:task_basket_shelf} complement
Fig.~\ref{fig:evaluation-overview} with the other three evaluated tasks.
Each figure follows the same row order: Robot Data in the Robot-ID condition,
converted KODA Data, and successful SONIC deployment in the Robot-OOD condition.
Four keyframes show the action progression within each row. Timestamps are
relative to the corresponding source clip; different rows are independent
recordings and are not temporally synchronized. The KODA row retains the human
view, synthesized robot view, and offline kinematic replay described in
Appendix~\ref{app:qualitative}. Bystanders are cropped out or pixelated.

\clearpage

\end{document}